\documentclass[preprint]{elsarticle}

\usepackage{hyperref}
\usepackage{url}
\usepackage{verbatim}
\usepackage{algorithm}
\usepackage{algorithmic}
\usepackage{amsmath}
\usepackage{amssymb}
\usepackage[toc,page]{appendix}
\usepackage{multicol}
\newtheorem{defn}{Definition}
\newtheorem{thm}{Theorem}
\newtheorem{lemma}[thm]{Lemma}
\newtheorem{corollary}{Corollary}
\newdefinition{rmk}{Remark}
\newproof{pf}{Proof}
\newcommand{\Ima}[1]{\textnormal{im}(#1)}
\newcommand{\Ker}[1]{\ker(#1)}
\newcommand{\Dom}[1]{\textnormal{dom}(#1)}
\newcommand{\Top}[1]{\ensuremath{(#1,\tau_{#1})}}

\usepackage{tikz-cd}

\begin{document}

\title{Coarse-Refinement Dilemma: On Generalization Bounds for Data Clustering\tnoteref{t1,t2}}

\author[1]{Yule Vaz\corref{cor1}}\ead{yule.vaz@usp.br}
\author[1]{Rodrigo Fernandes de Mello}\ead{mello@icmc.usp.br}
\author[1]{Carlos Henrique Grossi}\ead{grossi@icmc.usp.br}
\cortext[cor1]{Corresponding author.}
\address[1]{Institute of Mathematical and Computer Sciences, University of S\~ao Paulo, Trabalhador S\~{a}o Carlense 400, S\~{a}o Carlos, SP, Brazil}

\begin{abstract}
The Data Clustering (DC) problem is of central importance for the area of Machine Learning (ML), given its usefulness to represent data structural similarities from input spaces. Differently from Supervised Machine Learning (SML), which relies on the theoretical frameworks of the Statistical Learning Theory (SLT) and the Algorithm Stability (AS), DC has scarce literature on general-purpose learning guarantees, affecting conclusive remarks on how those algorithms should be designed as well as on the validity of their results. In this context, this manuscript introduces a new concept, based on multidimensional persistent homology, to analyze the conditions on which a clustering model is capable of generalizing data. As a first step, we propose a more general definition of DC problem by relying on Topological Spaces, instead of metric ones as typically approached in the literature. From that, we show that the DC problem presents an analogous dilemma to the Bias-Variance one, which is here referred to as the {\it Coarse-Refinement\/} (CR) dilemma. CR is intended to clarify the contrast between: (i) highly-refined partitions and the clustering instability (overfitting); and (ii) over-coarse partitions and the lack of representativeness (underfitting); consequently, the CR dilemma suggests the need of a relaxation of Kleinberg's richness axiom. Experimental results were used to illustrate that multidimensional persistent homology support the measurement of divergences among DC models, leading to a consistency criterion.
\end{abstract}

\begin{keyword}
  Data Clustering \sep Topology \sep Persistent Homology \sep Multidimensional Persistence \sep Algorithm Stability 
\end{keyword}

\maketitle

\section{Introduction}

Machine Learning (ML) is among the most important concepts to be considered while designing real-world applications from different domains~\citep{Zhao:2003,larkshmanan15,Zhan:2018,Tang:2018,Bablani:2019} by mainly relying on two paradigms: (i) the Supervised Machine Learning (SML), and (ii) the Unsupervised Machine Learning (UML). SML counts on fundamental proofs provided by the Statistical Learning Theory (SLT)~\citep{Vapnik95,luxburg09} while estimating a classification/regression function in form $f : X \rightarrow Y$, given an input space $X$ and class labels in an output space $Y$. All those proofs are driven to ensure that the empirical risk probabilistically converges to its expected value, so that it can be used to assess multiple learning models. This SLT framework cannot be employed to formulate or ensure learning in the context of UML, once class labels are not available but only inputs $x \in X$.  In this sense, UML attempts to represent data spatial structures according to the features composing $X$, being Data Clustering (DC) the most iconic approach of such branch. As matter of fact, some specific proofs have been already formulated~\citep{kleinberg02,bousquet02,ben-david08,ackerman10,carlsson10}, although UML still requires advances in order to have a stronger, preferably general, theoretical foundation to ensure learning.

As a step in such direction, the concept of Algorithmic Stability (AS)~\citep{bousquet02} supports learning guarantees in terms of bounded perturbations on the domain of a measurable function of random variables, even in the absence of labeled data~\citep{kutin02,poggio04}. In~\citep{mello09}, the Algorithm Stability was employed in an attempt to characterize learning on unsupervised online data, which resulted in the development of a method for concept drift detection. We then consider that Algorithmic Stability is an appropriate framework to study Data Clustering problems.

Considering clustering partitioning,~\citet{kleinberg02} formalizes the DC problem according to three necessary properties: (i) scale-invariance -- the partitions formed by a clustering algorithm should not depend on the distance scale among elements; (ii) consistency -- the partitions must not change whenever intra-cluster distances decrease and inter-cluster distances increase; and (iii) richness -- a clustering algorithm should be capable of producing all possible partitions for a distance function. \citet{kleinberg02} proved those properties cannot be simultaneously satisfied though in attempt to unify intuitive clustering notions, therefore even those basic axiomatic framework statements need some sort of relaxation to perform clustering. 

Based on ~\citet{kleinberg02} results,~\citet{ben-david08} propose the Clustering Quality Measure which guarantee the satisfiability for all~\citet{kleinberg02}'s axioms simultaneously. Although, we believe that richness is actually not mandatory, given its impacts on the algorithm stability as discussed in Sections~\ref{sec:generalizingDC} and~\ref{sec:onlearning}. Roughly speaking, richness imposes that some irrelevant and unstable partitions must be also produced by clustering algorithms, something that may not be desirable. In addition,~\citet{ackerman10} develop a taxonomy scheme for clustering properties from which we adopt the isomorphic invariance between clustering models.

Assuming that Hierarchical Clustering (HC) relaxes Kleinberg's axioms,~\citet{carlsson10} designed a theoretical framework to ensure data-permutation stability by taking advantage of ultrametric spaces built upon HC algorithms. The authors firstly proved that, after some modifications, the Single-Linkage (SL) agglomerative criterion is enough to ensure the same clustering model (dendrogram) for all input permutations $x \in \mathcal{X}$, and secondly confirmed the same result for different inputs following the same data distribution.

From those two main papers, this manuscript firstly consolidates those concepts in the sense of providing a general description for the Data Clustering (DC) problem by using topological spaces, thus complementing~\citet{carlsson10}'s study that assumes data in some metric space. Secondly, we discuss on the practical usefulness of~\citet{kleinberg02}'s richness property given its impacts on the clustering algorithm consistency, as later discussed in Sections~\ref{sec:generalizingDC} and~\ref{sec:onlearning}. Roughly speaking, richness imposes that either irrelevant or unstable partitions must be also produced which are not desirable. Finally, we conclude that over-refined or over-coarse HC partitions tend to be either unstable or irrelevant when data is subject to bounded perturbations, something associated to the Bias-Variance dilemma~\citep{Geman92} in terms of the space of the admissible partitions, thus suggesting that Kleinberg's richness should be anyway relaxed.

To complement~\citet{carlsson10}'s study, this paper shows that topological spaces sufficiently model the DC problem, allowing to derive consistency results to ensure clustering generalization from a more general point of view (Sections~\ref{sec:generalizingDC} and~\ref{sec:onlearning}). This consistency considers topological features primarily associated with connected components, holes and voids~\citep{munkres2000}, which cannot be directly represented once the underlying topological space is unknown. However, given we have access to some points cloud, the consistency of topological structures can be assessed by evaluating isomorphisms between homology groups (homology equivalences)~\citep{carlsson09}.

Equivalent spaces from the homological perspective may not be homeomorphic~\citep{hatcher2002}. Even though, homology equivalence preserves holes, voids and connected components of geometric objects, which are defined by homology classes, i.e., by elements of the homology group. In this particular context, we claim that inferior (fine-grained, e.g. at the bottom level of a dendrogram) hierarchies in some HC model are not consistent for homology classes as data points are subject to data inclusion. In this scenario, persistent homology~\citep{Edelsbrunner2000} is suitable to study the homology groups extracted from the hierarchies of a HC model as it allows to analyze changes in the number of connected components and voids, what is formally defined in terms of inclusions $\Top{X_i} \subseteq \Top{X_j}$ of the corresponding topological spaces. Hence, our goal is to verify how persistent homology is affected after the acquisition of new data so that we find the collection of hierarchies satisfying generalization bounds for the HC problem.

That motivated us to study the stability proofs by~\citet{Cohen07,Chazal09} of persistence diagrams. Such diagrams describe how homology classes change throughout the sequence of topological spaces inclusions $\Top{X_1} \subseteq \Top{X_2} \subseteq \dots \subseteq \Top{X_n}$ which, in the context of this paper, are associated to the hierarchies of a given HC model (or dendrogram). Although the relevant contributions on the stability of persistence diagrams by~\citet{Cohen07,Chazal09}, their approaches did not considered variations on the topological spaces produced by new data. Such data insertions required us to employ the concept of multifiltration (bifiltration specifically), introduced in~\citep{carlsson2009multi}, in order to represent persistent homology along data inclusions~\citep{ville1939}.

In summary, the {\bf main contributions} of this paper are:
i) the conclusion that Kleinberg's richness property~\citep{kleinberg02} may lead to inconsistent partitions;
ii) the conclusion that Topological Spaces produce relevant features for DC consistency analysis;
iii) a new DC problem formulation based on topological spaces;
iv) a new HC problem formulation based on topological spaces;
v) the formulation of the Coarse-Refinement dilemma based on homology groups which is associated with the Bias-Variance dilemma from supervised learning~\citep{Vapnik95};
vi) the formulation of generalization bounds for homology groups in clustering structures, based on the Coarse-Refinement dilemma;
vii) the proof that our proposed DC generalization bound is an upper limit for~\citet{carlsson10}'s consistency;
viii) besides the theoretical contributions, we show experimental results to confirm that over-refined clusters produce inconsistent homology groups.

This paper is organized as follows: Section~\ref{sec:related_work} briefly introduces some studies related to the formalization of theoretical frameworks in the context of the Data Clustering (DC) problem; Section~\ref{sec:generalizingDC} introduces a general formulation for the DC and HC problems; Section~\ref{sec:bias-variance} discusses the Coarse-Refinement Dilemma considering the homology group $H_0$; Section~\ref{sec:onlearning} shows that homology groups of degree greater than zero are affected by over-refined and over-coarsed topologies; 
Section~\ref{sec:carlsson} compares our proposed generalization bounds to~\citet{carlsson10}'s consistency;
finally, conclusions and future directions are provided in Section~\ref{sec:conclusion}.

\section{Related work}\label{sec:related_work}

Data Clustering (DC) faces many challenges in defining and guaranteeing generalization from datasets, as it does not rely on labels and, consequently, it cannot take advantage of computing any evident error measurement such as risk~\citep{luxburg09}. While studying this issue, \citet{kleinberg02} considered that a clustering model is an application of a mapping $f$ on top of a distance function $d:I\times I \rightarrow \mathbb{R}^{+}$, given $I$ contains indices of data points in some fixed-size set $S$, disregarding its ambient space though~\citep{ambientspace}. From this initial setup,~\citet{kleinberg02} defined three properties to be respected in order to assess clustering algorithms and models:
\begin{itemize}
    \item Scale-invariance: Given a distance and a clustering function, $d$ and $f$, and a scalar $\alpha$, the following must hold $f(d) = f(\alpha d)$. Thus, the similarity representation over $S$ must be consistent with the units of measurement;
    \item Consistency: Let $\Gamma$ be a partition of $S$ and $d, d'$ two distance functions. Function $d'$ is referred to as a $\Gamma-$transformation of $d$ if: (i) for all $i,j \in S$ belonging to the same cluster, $d'(i,j) \leq d(i,j)$; and (ii) for all $i,j \in S$ belonging to different clusters, $d'(i,j) \geq d(i,j)$. Consistency holds if $f(d') = f(d)$ whenever $d'$ is a $\Sigma-$transformation of $d$. Intuitively, this property is assured when the partition is maintained whenever intra-cluster distances reduce and inter-cluster distances increase;
    \item Richness: Let $\text{Range}(f)$ be the set of all partitions $\Gamma$, given some distance function $d$ such that $f(d) = \Gamma$. The richness property is guaranteed if and only if $\text{Range}(f)$ is equivalent to the set of all partitions of $S$. In other words, for any partition $\Gamma_i$ of $S$, there exists some $d_i$ such that $f(d_i) = \Gamma_i$.
\end{itemize}

The author also proves that those three properties cannot be simultaneously ensured, and claim that one of them is always somehow relaxed to obtain a clustering model. Nonetheless, if we consider $f$ as a statistical model, richness does not appear to be suitable for the Data Clustering scenario as, in the spirit of the Bias-Variance dilemma, it would permit the production of biased models, thus leading to phenomena similar to under (single cluster) and overfitting (every point is a cluster).

In addition,~\citet{carlsson10} relax Kleinberg's properties to prove the stability and consistency for their adapted single-linkage HC strategy. They assume hierarchical methods on metric spaces $(X,d)$ to build up dendrograms, i.e., structures that map the groups of an HC model into the real line, associating every cluster with the required radii to form it. They also show dendrograms are equivalent to ultrametric spaces, allowing them to compare HC models using the Gromov-Hausdorff distance~\citep{gromov1981}, confirming their adapted single-linkage strategy is the only one in the class of linkage algorithms (uniqueness) to be stable and consistent according to the metric space. More precisely, they conclude that two different ultrametric spaces built from identically and independently distributed samples from the same data distribution give rise to isometric spaces as the sample size goes to infinity. 

Complementary to those relevant studies, topological features could be used to derive other theoretical results for the DC and HC problems. In that sense, persistent homology is particularly useful as it describes when homology classes appear or vanish throughout a sequence of topological inclusions in attempt of representing hierarchical clusterings~\citep{Edelsbrunner2002}. The persistence diagram is typically employed to represent the birth and death of homology classes, whose stability was already proven by considering changes in tame functions~\citep{Cohen07}. Taking advantage of such foundation to prove stability,~\citet{Cohen07} assumed that: (i) the topological space must be triangulable; (ii) the topological space must be fixed; and (iii) tame functions must be continuous. This has also motivated~\citet{Chazal09} to prove that persistence diagrams are stable when topological spaces are not triangulable nor tame functions are continuous, yet topological spaces must be fixed. In our scenario, neighborhoods may change when data is subject to perturbations, so we cannot assume the topology to be fixed. Therefore, the main goal of this paper is to study the consistency in the presence of adaptable topological spaces, guaranteeing the property of isomorphism invariance as in~\citep{ackerman10}.

Hence we adopted the concept of multifiltration (bifiltration precisely), introduced in~\citep{carlsson2009multi}, from which the persistence of homology groups can be studied along variations of two or more parameters. For instance, a bifiltration can be used to study the radius $r$ and the density $\rho$ of the DBSCAN algorithm~\citet{Ester96} to obtain adequate clusters given a target application. In our work we consider the inclusion of new data points as a second dimension in the bifiltration such that homology classes are associated with a pair $(r,X^m)$ in which $X^m$ is a dataset with the inclusion of $m$ new samples and $r$ is a variable related to the level of the HC (e.g. an acceptable radius). In order to study the persistent homology of such bifiltration and its $p$-th Betti-numbers consistency, the DC and HC problems must be formulated in terms of topological spaces, as shown in the next section.

\section{Generalizing the Data Clustering problem}\label{sec:generalizingDC}

The Data Clustering (DC) problem typically relies on metric spaces in order to describe similarities among dataset instances, unfortunately losing the inherent abstraction that it could take advantage from topological spaces. As discussed in this section, the DC problem should be represented in a more general mathematical space so it can be restricted when and if necessary, according to the target learning task. For example, if no restriction is needed, one can analyze topological features to study the data space and understand more general stability/consistency criteria on top of structures such as holes, voids and connected components~\citep{carlsson10}. However, if some restriction is part of the target task, one can still endow such space with some topology and take advantage of the same general stability/consistency criteria, as formulated in this manuscript. 

In this context, we consider data points are acquired from some unknown topological space $\Top{Z}$, whose modeling is the goal of data clustering. To produce such model, the DC problem assesses the similarities among dataset instances in order to organize them into clusters, each one corresponding to a set of neighborhoods of data elements. A neighborhood of $x \in \Top{X} \subset \Top{Z}$, having $\Top{X}$ as the resultant topological space sampled from $\Top{Z}$, defines an open set of $X$ necessarily containing $x$.

\begin{rmk}\label{rmk:neigh_topology}
The goal of Data Clustering is to approximate topological features of an unknown topology\/ $\tau_Z$ on a space\/ $\Top{Z}$ from which data is sampled. In practice, a subspace of a larger topological space\/ $\Top{\Omega}$ {\rm(}typically, a closed cube in $\mathbb{R}^p${\rm)}. From the data instances\/ $x\in X\subset Z$, one attempts to approximate topological features of $\Top{Z}$ by means of a {\rm neighborhood map\/} $\eta:x\mapsto \mathcal{N}(x) \subset \tau_{\Omega}$, in such a way that, ideally, $\Top{Z}$ and\/ $(X,\mathcal{N}(X))$ are homeomorphic. Here, $\mathcal{N}(x)$ is an open neighborhood of $x$ and $\mathcal{N}(X):=\overline{\bigcup_{x\in X}\mathcal{N}(x)}$ is endowed with the subspace topology\/ {\rm(}we denote by $\overline A$ the closure of the topology $\tau_{A}${\rm).} We call the pair $(X,\mathcal{N}(X))$ a {\rm neighborhood topological space}.
\end{rmk}

For the sake of illustration, the DBSCAN algorithm~\citep{Ester96} employs a neighborhood map that uses open balls around every data instance in some metric space, considering the density of points in every neighborhood is enough to form a new cluster. Observe this concept differs from ours in which all open sets are taken into account.

Neighborhood topologies induce equivalence relations on $N(X):=\bigcup_{x\in X}\mathcal{N}_x$ (an example will be given below) which determine the pertinence of every data instance to a given cluster. For example, the single-linkage algorithm induces an equivalence relation $\sim_r$, referred to as $r$-relation by~\cite{carlsson10}, in which a metric space $(X,d)$ is built from some dataset $X$ using a distance function $d:(X,d) \rightarrow \mathbb{R}$, so that $x \sim_r x'$ holds when there is a set $\{x_1,x_2,\dots,x_n\}$ such that $d(x,x_1) < r,d(x_1,x_2) < r,\dots,d(x_n,x') < r$.

We assume Data Clustering models to be built upon independently and identically sampled instances from which some topological space $\Top{X} \subset \Top{Z}$ is obtained, so that there is a probability measure $P$ over some unknown underlying topological space $\Top{Z}$. In order to ensure such assumption, the set $Z$ must also endows a probability space. In this scenario, we need to endow the Borel $\sigma$-algebra of the larger topological space $\Top{\Omega}$ (see Remark \ref{rmk:neigh_topology}) with a probability measure that is supported on $Z$. 

Summarizing, in this paper, we define the Data Clustering (DC) problem as follows.
\begin{defn}[Data Clustering problem]
The Data Clustering problem consists in finding an adequate neighborhood topology $\mathcal{N}(X)$ of $(X,\mathcal{N}(X))$ {\rm(}see\/ {\rm Remark\/ \ref{rmk:neigh_topology})}, where $x \in X$ and $\Top{X} \subset \Top{Z} \subset \Top{\Omega}$ is a known topological space sampled from the unknown underlying topological space $\Top{Z}$. The neighborhood topology $N$ should approximate topological features of the unknown topology $\tau_Z$. Random variables are independently and identically sampled from some unknown probability distribution which is supported on $Z$. Clusters are obtained from an equivalence relation derived from $\mathcal{N}(X)$.
\end{defn}
As claimed in~\citep{carlsson09}, the use of hierarchical schemes is more informative than choosing a single neighborhood topology. The Hierarchical Clustering (HC) problem considered in this paper is defined as follows.
\begin{defn}[Hierarchical Clustering problem]
The Hierarchical Clustering problem consists in finding, for each $i\in\{1,\dots,n\}$ in which $n$ is a refinement index (e.g. such as the radius index for a metric space), an adequate neighborhood topology $\eta_i:x\mapsto \mathcal{N}_i(x)$ as in the previous definition. Moreover, we require that the neighborhood topology $\mathcal{N}_i(X):=\overline{\bigcup_{x \in X} \mathcal{N}_i(X)}$ determined by $\eta_i$ is contained in $\mathcal{N}_j(X)$ whenever $i\leq j$.
\end{defn}

In this context, the comparison among neighborhood topologies produced from different datasets acquired by the same i.i.d. distribution $P(Z)$ can be performed by verifying homeomorphisms among them. Two problems occur in this sense, though:
\begin{itemize}
	\item As the underlying topological space $\Top{Z}$ is unknown, it is actually impossible to verify when a neighborhood topology is homeomorphic to $\tau_Z$; this avoids the possibility of a direct design of an adequate neighborhood topology;
    \item Consider two subsets $X,X'\subset Z$ and the respective neighborhood topologies $\mathcal{N}(X)$ and $\mathcal{N}(X')$ (see Remark \ref{rmk:neigh_topology}). Since the probability distribution $P$ is unknown, homeomorphisms between $(X,\mathcal{N}(X))$ and $(X',\mathcal{N}(X'))$ cannot be explicitly studied.
\end{itemize}

Then, some notion of ``approximately'' equal topologies is required in order to analyze whether a (hierarchical) clustering algorithm is capable of consistently grouping data. To this end, we use homology: in some sense, homology allows one to compare some features of topological spaces by mapping topological structures into algebraic ones~\citep{hatcher2002}.

In this paper, we are mainly interested in {\it singular\/} homology. This particular kind of homology uses the continuous images of standard {\it simplices\/} into a topological space $\Top{X}$ in order to ``get a sense'' of the topology $\tau_X$. More precisely, the standard $p$-simplex is the convex hull of the $n+1$ standard unit vectors in $\mathbb{R}^{p+1}$, that is, $\Delta^p:=\{x\in\mathbb{R}^{p+1}\mid\sum_{i=0}^{p} x_i = 1, x_i \geq 0\}$ and a singular $p$-simplex in $(X,\tau_X)$ is a continuous map $\sigma:\Delta^p \to (X,\tau_X)$. The boundary of the standard $p$-simplex is made up of $(p-1)$-simplices; so, we define the boundary $\partial \sigma$ of the singular $p$-simplex $\sigma$ as being the formal sum of the singular $(p-1)$-simplices that are given by restricting $\sigma$ to the faces of $\Delta^p$. In this formal sum, we alternate signs so orientation is taken into account.

The homology groups can now be constructed as follows. First, one takes the free abelian group $C_p$ generated by all singular $p$-simplices on $\Top{X}$, i.e., the formal finite sums of singular $p$-simplices with integer coefficients. Elements of $C_p$ are called {\it singular $p$-chains}. The boundary operator $\partial$ immediately extends to an operator $\partial_p:C_p\to C_{p-1}$ giving rise to the {\it chain complex}
\[
\centering
\begin{tikzcd}
\dots \arrow[r,"\partial_{p+1}"] & C_{p} \arrow[r,"\partial_p"] & C_{p-1} \arrow[r,"\partial_{p-1}"] & \dots \arrow[r,"\partial_2"] & C_{1} \arrow[r,"\partial_{1}"] & C_0 \arrow[r,"\partial_0"] & 0.
\end{tikzcd}
\]
The image of the boundary operator $\partial_p$ provides the $p$-{\it boundaries\/} while its kernel provides the $p$-{\it cycles.} It is not difficult to show~\citep{hatcher2002} that the composition $\partial_p\circ\partial_{p+1}=0$. So, $\Ima{\partial_{p+1}}\subset\Ker{\partial_p}$ and the $p$-th homology group of the topological space $\Top{X}$ is defined as the quotient
\begin{equation}
    H_p[\Top{X}] := \Ker{\partial_p}/\Ima{\partial_{p+1}}.
\end{equation}
The inclusion $\Ima{\partial_{p+1}}\subset\Ker{\partial_p}$ says that the ``boundary of a boundary'' is trivially empty. So, homology groups detect the cycles in $C_p$ whose boundaries are empty but not by the trivial reason of already being a boundary (of a higher dimensional singular simplex) themselves. The presence of a cycle in $\Top{X}$ that is not the boundary of a higher dimension singular simplex can be seen as the presence of an $n$-dimensional void (or hole) in $\Top{X}$.

The $n$-dimensional voids in $\Top{X}$ are characterized by its $p$-th Betti-number which is, by definition, the rank of the abelian group $H_p[\Top{X}]$. One can also define $p$-th Betti-numbers as the dimension of the vector space $H_p(X,\mathbb Q)$; the definition of $H_p(X,\mathbb Q)$ is essentially the same as that of $H_p[\Top{X}]$ but we take formal sums of singular simplices with coefficients in the field $\mathbb Q$ of rational numbers -- a subtle difference allowing one to endow $H_p(X,\mathbb Q)$ with a linear structure. For example, for a topological space $T$ built up from a two-dimensional torus, $H_0(T) = \mathbb{Z}$, $H_1(T) = \mathbb{Z}^2$ and $H_2(T) = \mathbb{Z}$. So, the corresponding $p$-th Betti-numbers are $\beta_0 = 1$, $\beta_1 = 2$, and $\beta_2 = 1$.

In these settings, topological spaces built up from datasets can be compared in terms of their homology groups without the introduction of a function to represent perturbations on such sets. More precisely, given a dataset $X$ and a perturbed version $X^m$ which consists of $X$ with $m$ added points, one expects that the probability of $H_p[\Top{X}]$ and $H_p[\Top{X^m}]$ not being isomorphic tends to zero as $m\to\infty$ ({\it consistency\/}).

\section{Coarse-Refinement dilemma of the homology group $H_0$}\label{sec:bias-variance}

In the previous section, we argued that isomorphisms at the level of (singular) homology groups can be useful to compare topological spaces constructed from some Data Clustering algorithm. Homology groups encode how many connected components, holes, or voids (topological structures) are present in a topological space. If topological spaces $\Top{X},\Top{X'}$ have the same topological structure, i.e., if they are homeomorphic, their homology groups $H_p[\Top{X}]$ and $H_p(\Top{X'})$ are isomorphic. The converse is not true.

In spite of that, homology is used in Topological Data Analysis (TDA) to study the topology of point clouds, because it does not require any assumption on the unknown underlying topology $\tau_{Z}$~\citep{carlsson09,Chazal17}. In fact, neighborhood topologies (see Remark~\ref{rmk:neigh_topology}), typically based on metric spaces~\citep{Ester96,carlsson10}, are used in TDA to extract topological features from point clouds. A more general analysis can be performed by considering arbitrary open neighborhoods instead of metric ones. In addition, the homology groups can be used to compare how a neighborhood topology $\mathcal{N}(X)$ changes when compared to a perturbed version $\mathcal{N}(X')$ of itself, analyzing the stability of $(X,\mathcal{N}(X))$ by comparing $H_p[(X,\mathcal{N}(X))]$ with its expected homology $\mathbb{E}\{H_p[(X,\mathcal{N}(X))] \}$.

If $\mathcal{N}(X)$ is refined enough to represent each element into a single cluster (singletons), a problem would occur at the level of topological spaces: if $|\mathcal{N}(X)| \neq |\mathcal{N}(X')|$, then such spaces are not homeomorphic as their number of connected components differ, hence their $H_0$-homological groups are not isomorphic; so $|\mathcal{N}(X)| = |\mathcal{N}(X')|$ should hold when over-refined spaces are considered. However, no assumption about the cardinality of $\tau_Z$ can be given since $\Top{Z}$ is unknown and, therefore, $|\mathcal{N}(X)| = |\mathcal{N}(Z)|$ may not hold.
\begin{thm}
    Two over-refined topologies $\mathcal{N}(X)$ and $\mathcal{N}(X')$ produces isomorphic $0$-dimensional homology groups iff $|\mathcal{N}(X)| \neq |\mathcal{N}(X')|$.
\end{thm}

\begin{pf}
Assume that, given a neighborhood topology $\mathcal{N}(X)$, the space $\Top{\Omega}$ (we remind the reader that, typically, $\Top{\Omega}$ is a closed cube in $\mathbb{R}^p$) is endowed with a probability measure $P$ (in the Borel $\sigma$-algebra of $\Top{\Omega}$) supported in $\Top{Z}\subset\Top{\Omega}$ and such that each corresponding cluster has measure $\leq\epsilon$. Consider the presence of some sampled perturbation $x'=x+\delta$, where $\delta$ is an element of a measurable set $D$ endowed with a probability function. One issue that motivated the dilemma on over-coarse versus over-refined topologies (in the same sense of the Bias-Variance dilemma) in this paper is that whenever an element in any $C \in \mathcal{N}(X)$ becomes unrelated with any neighborhood in $\mathcal{N}(X)$, the topology changes as well as its homology group $H_0$. Then, we formulate the probability of the topology being ``cut'' (or divided) as
\begin{equation}
\begin{array}{rcl}
P((x+D) - \bigcup_{C \in \mathcal{N}(X)} C) & = & P(x+D) \\ 
                                            & & - P\left(\bigcup_{C \in \mathcal{N}(X)}(C \cap(x+D))\right) \\
                                            & = & 1 - \sum_{C \in \mathcal{N}(X)} P(C|x+D).
\end{array}
\end{equation}
Note that, if $\epsilon \rightarrow 0$, i.e., tends to over-refinement, and some measure $\mu_Z(D)$ does not depend on $\epsilon$, then $P(C|x+D) \rightarrow 0$ and $P(x \notin C) \rightarrow 1$ for every $C\in\mathcal{N}(X)$. In this case, the topological space determined by $\mathcal{N}(X')$ is not homeomorphic to the one determined by $\mathcal{N}(X)$ nor its homology group $H_0[(X',\mathcal{N}(X'))]$ is isomorphic to $H_0[(X,\mathcal{N}(X))]$. On the other hand, if the elements $C\in\mathcal{N}(X)$ cover $x+D$, then $P(x \notin C) = 0$ for every $C\in\mathcal{N}(X)$ and, therefore, $H_0[(X',\mathcal{N}(X'))] \cong H_0[(X,\mathcal{N}(X))]$.
\end{pf}

The neighborhood topology $\mathcal{N}(X)$ could produce an over-coarse cluster that consists only of the universal set $\Omega$, giving rise to homeomorphic spaces for every perturbation of $\Top{X}$, but overlooking homology classes with degree greater than one could appear if the topology were refined. Thus, we conclude that there is a trade-off between probabilistic consistence and data representation, such as depicted in the Bias-Variance dilemma~\citep{mello18}. This leads to the same common issue resultant from the SLT, in which a SML algorithm should never produce all possible classification functions given that would lead to overfitting~\citep{Vapnik95}. In the Data Clustering context, a clustering algorithm must not produce all possible neighborhood topologies, because this would lead to unstable clustering related to over-refined topologies.

\section{Generalization bounds for homology groups}\label{sec:onlearning}

The trade-off resultant of the refinement of $\mathcal{N}(X)$ can be studied using persistent homology, from the most refined topological space to the coarsest one. In this sense, we have inclusions of topological spaces $\mathcal{F}(X,\eta):=(X,\mathcal{N}_1(X))\subseteq (X,\mathcal{N}_2(X))\subseteq\dots\subseteq (X,\mathcal{N}_n(X))$, known as a {\it filtration}, that corresponds to the levels of a hierarchical clustering (here, $\mathcal{N}_i(X)=\overline{\bigcup_{x\in X}\eta_i(x)}$ stands for the neighborhood topology from which arises the topological space $(X,\mathcal{N}_i(X))$ (see Remark~\ref{rmk:neigh_topology}). Our proposal is to find suitable neighborhood topologies $\mathcal{N}_i(X)$ resultant from hierarchical clusterings that consistently, as $i$ increases, represent the persistence of homology classes even when data is subject to perturbations. 

Along the filtration, new simplices may fill/create new cycles in \break$H_p[(X,\mathcal{N}_i(X))]$ or merge/create connected components accordingly to a tame function:
\begin{defn}[Tame functions]\label{def:tame}
	Let $\Top{X}$ be a topological space. A tame function is a continuous map $f:\Top{X} \rightarrow \mathbb{R}$ such that $\Top{X}_{t_i} \subseteq \Top{X}_{t_j}$ whenever $t_i<t_j$, where $\Top{X}_t:=f^{-1}(-\infty,t]$ is taken with the subspace topology. Moreover, a tame function $f$ must satisfy the following properties:
	\begin{itemize}
		\item The homology groups $H_p(\Top{X}_t)$ are of finite rank for every $p$;
		\item There are finitely many $t_i\in\mathbb{R}$ such that $H[\Top{X}_{t_i}]$ and $H[\Top{X}_{t_i+\epsilon}]$ are not isomorphic; in which $t_i$'s are called the\/ {\rm critical values} of $f$.
	\end{itemize}
\end{defn}
Whenever a homology class emerges at $H_p[(X,\mathcal{N}_j(X))]$ and vanishes at \break $H_p[(X,\mathcal{N}_j(X))]$, given $i<j$, its persistence is defined as $t_j-t_i$. Note that this process is a generalization for the Morse function~\citep{morse1947}.

Roughly speaking, given $t_{i-1},t_i\in\mathbb{R}$, a persistent homology group identifies the homology classes which are present along the interval $[t_{i-1}$, $t_{i})$ of the co-domain of the tame function. More precisely, the persistent homology groups are defined as follows.
\begin{defn}[Persistent Homology Group]
    Let\/ $\Top{X}$ be a topological space equipped with the filtration that arises from a tame function\/ $f$, as in the previous definition. Given\/ $t_{i-1}<t_i$, we have the inclusion $f^{i,j}:\Top{X}_{t_i} \subseteq \Top{X}_{t_j}$. The persistent homology group of degree\/ $p$ is the image of the induced homomorphism
	\begin{equation}
		\mathbf{f}^{i,j}_p : H_p[\Top{X}_{t_{i}}] \rightarrow H_p[\Top{X}_{t_{j}}].
	\end{equation}
\end{defn}
The rank of the image of $\mathbf{f}^{i,j}_p$ is called the $(i,j)$-{\it persistent Betti-number,} i.e.,\break $\beta^{i,j}_p = \text{rank}\,\Ima{\mathbf{f}^{i,j}_p}$. It allows us to compute the number of homology classes persisting in an interval $[i,j)$. In the context of Hierarchical Clustering, we consider a model to be adequate whenever the corresponding homology classes are likely to persist. We claim that over-refined topologies do not provide homology classes that tend to persist when data is subject to perturbations given by instance inclusions.  

In order to prove that homological groups are not stable when it comes to over-refined topologies, let us consider perturbations on some dataset $X=\{x_1,x_2,\dots,x_n\}$ after the inclusion of $m$ new instances such that \break $X^{m}=\{x_1,x_2,\dots,x_n,x'_1,\dots,x'_m\}$. Let $\mathcal{N}(X)$ be a neighborhood topology with the corresponding topological space $\mathcal{X}$ and assume that there exists a tame function $f:(X,\mathcal{N}(X))\to\mathbb{R}$ giving rise to the filtration $\mathcal{F}(X,\eta)$ as in Definition~\ref{def:tame}. New data must be identically and independently sampled from the unknown underlying topology according to an unknown probability distribution and it produces a new neighborhood topology $\mathcal{N}_i(X^m)$ of $X^m$ such that if its $p$-dimensional homology group is not isomorphic to that of $\mathcal{N}_i(X)$, then the clusters in $\mathcal{N}_i(X)$ are not consistent according to these homology groups ($p$-homology consistency). For simplicity and without loss of generality, from now on we will define the topological spaces produced from neighborhood topologies $\mathcal{N}_i(X)$ and $\mathcal{N}_i(X^m)$ as, respectively $\mathcal{X}_i$ and $\mathcal{X}_i^m$.

The impact of acquiring a new sample can be analyzed considering the coarse-grain of the corresponding topological space through the sampling of a new element $x'$, forming, then, inclusions $\iota_X := \mathcal{X}_i^{l < q} \subseteq \mathcal{X}_i^{q}$ with $l,q < m$. This sequence of topological inclusions induces another filtration; assume that there exists some tame function $g$ such that $\mathcal{X}_i^{q} = g^{-1}(-\infty,q]$. The inclusions $\iota_\mathcal{N} := \mathcal{X}_i^q \subseteq \mathcal{X}_j^q$ and $\iota_X$ define a bifiltration~\citep{carlsson2009multi}, as follows:
\[
\centering
\begin{tikzcd}
\mathcal{X}_{1} \arrow[d,"\iota_X"]\arrow[r,"\iota_\mathcal{N}"] & \mathcal{X}_{2} \arrow[d,"\iota_X"]\arrow[r,"\iota_\mathcal{N}"] & \dots \arrow[d,"\iota_X"]\arrow[r,"\iota_\mathcal{N}"] & \mathcal{X}_{k}\arrow[d,"\iota_X"] \\
\mathcal{X}_{1}^1 \arrow[d,"\iota_X"] \arrow[r,"\iota_\mathcal{N}"] & \mathcal{X}_{2}^1 \arrow[d,"\iota_X"] \arrow[r,"\iota_\mathcal{N}"] & \dots \arrow[d,"\iota_X"] \arrow[r,"\iota_\mathcal{N}"] & \mathcal{X}_{k}^1 \arrow[d,"\iota_X"]\\
\vdots \arrow[d,"\iota_X"] \arrow[r,"\iota_\mathcal{N}"] & \vdots  \arrow[d,"\iota_X"] \arrow[r,"\iota_\mathcal{N}"] & \vdots  \arrow[d,"\iota_X"] \arrow[r,"\iota_\mathcal{N}"] & \vdots \arrow[d,"\iota_X"] \\
\mathcal{X}_{1}^m \arrow[r,"\iota_\mathcal{N}"]& \mathcal{X}_{2}^m \arrow[r,"\iota_\mathcal{N}"] & \dots \arrow[r,"\iota_\mathcal{N}"] & \mathcal{X}_{k}^m
\label{diag:bicomp}
\end{tikzcd}
\]
such that, for every $i,j,X^l$ and $X^q$, with $q > l$, the following diagram is commutative:
\[
\centering
\begin{tikzcd}
\mathcal{X}_{i}^q \arrow[d,"\iota_X"]\arrow[r,"\iota_\mathcal{N}"] & \mathcal{X}_{j}^q \arrow[d,"\iota_X"]\\
\mathcal{X}_{i}^r  \arrow[r,"\iota_\mathcal{N}"] & \mathcal{X}_{j}^r
\end{tikzcd}
\]
In this work we consider that every simplicial complex of $\mathcal{X}^l_i$ and $\mathcal{X}^q_i$ in the filtrations, with $l,q=1,\dots,m$, must be associated with the same pre-image in $f$ in order to allow their comparison. In this sense, note that $i$ may not correspond to a critical point for all $\mathcal{X}_i^q$. For instance, consider the case of $H_0$ when a neighborhood topology $\mathcal{N}_i(X)$ is constructed using open balls and let $\mathcal{N}_i(X^m)$ be a perturbed neighborhood topology, where the radius associated to $i$ is $r_i$. If the diameter of $\mathcal{N}_i(X^m)$ is sufficiently greater than that of $\mathcal{N}_i(X)$, the probability that a new instance $x'_q$ forms another cluster is close to one because the added data is not likely to belong to an existing cluster thus having probability zero. This implies that $\text{rank}\{\Ima{\mathbf{f}^{0,i}_0 \circ \mathbf{g}^{0,m}_0}\} = \text{rank}\{\Ima{\mathbf{f}^{0,i}_0}\}+m$, where $\mathbf{g}$ is the map induced on homology by the data inclusions, being related to the tame function $g$. 

In this sense, $\mathbf{f}^{0,i}_0(H_p(\mathcal{X}_i))$ will never represent well $\mathbf{f}^{0,i}_p(H_p(\mathcal{X}_i^m))$ and the analyzed topological features will change as new data are included. In order to $\mathbf{f}^{0,i}_0(H_p(\mathcal{X}_i))$ represent well $\mathbf{f}^{0,i}_p(H_p(\mathcal{X}_i^m))$, $\mathbf{g}^{0,m}_p$ must has, almost certainly, to be an isomorphism. Therefore, a DC model can be considered consistent as follows:
\begin{defn}[DC $p$-homology consistence]
     A DC model associated with the neighborhood topology $\mathcal{N}_i(X)$, is $p$-homology consistent if, as $m \rightarrow \infty$, for all $q=1,\dots,m$, $\mathbf{g}^{0,q}_p(H_p(\mathcal{X}_i))$ is almost surely an isomorphism.
     \label{def:dc-p-hom-cons}
\end{defn}
Although, hierarchical structures require the study of a filtration also throughout the domain of the tame function $f$. In this sense, we have to consider a set of morphisms $\mathbf{g}^{0,m}_p$ applied over $\mathbf{f}^{0,k}_p(H_p(\mathcal{X}_k))$ for $k=i,\dots,j$ in order to define isomorphisms among them. In the ideal scenario, all $\mathbf{g}^{0,m}_p$ are isomorphic for $k=i,\dots,j$. But note that, in hierarchical clustering, the indices of the filtrations are previously chosen, hence the set of morphisms $\mathbf{g}^{0,m}_p$ is limited by this choice. Critical points existent in any interval $[k-1,k)$ will not be considered and, therefore, there is a resolution problem (in terms of $\Dom{f}$) inherent to this analysis. Although, assuming this limited set of $\mathbf{g}^{0,m}_p$, the behavior of $\mathbf{f}^{i,j}_p(H_p(\mathcal{X}_k))$ can be study as:
\begin{lemma}
Given the persistences $\mathbf{f}_p^{i,j}$ and $\mathbf{g}_p^{0,m}$ of a bifiltration, if $\mathbf{g}^{0,m}_p(H_p(\mathcal{X}_i))$ and $\mathbf{g}^{0,m}_p(H_p(\mathcal{X}_i^m))$ are isomorphisms then $\mathbf{f}_p^{0,i}(H_p(\mathcal{X}_0^m)) \cong \mathbf{f}_p^{0,i}(H_p(\mathcal{X}_0))$ and $\mathbf{f}_p^{i,j}(H_p(\mathcal{X}_i^m)) \cong \mathbf{f}_p^{i,j}(H_p(\mathcal{X}_i))$.
\label{lemma:isomorph-insertion}
\end{lemma}

\begin{pf}
    The proof follows from the commutativeness of the bifiltration diagram.
\end{pf}

Note that, considering Lemma~\ref{lemma:isomorph-insertion}, if isomorphism is almost certain for \break $\mathbf{g}^{0,q}_p(H_p(\mathcal{X}_i))$ and $\mathbf{g}^{0,q}_p(H_p(\mathcal{X}_j))$ for $q=1,\dots,m$ as $m \rightarrow \infty$, then, \break$\mathbf{f}_p^{i,j}(H_p(\mathcal{X}_i^m)) \cong \mathbf{f}_p^{i,j}(H_p(\mathcal{X}_i))$ is likely to occur for any included data, i.e., $\mathbf{f}_p^{i,j}(H_p(\mathcal{X}_i))$ consistently represent $\mathbf{f}_p^{i,j}(H_p(\mathcal{X}_i^m))$ for the values of $i,j$. In sequence, we define a $p$-homology as:
\begin{defn}[HC $p$-homology consistence]
     A HC model associated with the filtration $\mathcal{F}(X,\eta)$, is $p$-homology consistent if for all $i \leq k \leq j$, $\mathcal{N}_k$ is $p-$homology consistent (Definition~\ref{def:dc-p-hom-cons}).
     \label{def:hc-p-hom-cons}
\end{defn}

In addition, the evaluation if $\mathbf{g}^{0,m}_p(H_p(\mathcal{X}_i))$ is non-isomorphic not only allows to locally study $\mathcal{X}_i$ but also to represent $\mathbf{f}^{i,t}_p$ as $t\to\infty$. In order to prove that, consider the following theorem
\begin{thm}
    Given the associated simplicial complexes built up from $\mathcal{X}_i$, if $\mathbf{g}^{0,m}_p(H_p(\mathcal{X}_i))$ is not an isomorphism then $\mathbf{f}^{i,t}_p(H_p(\mathcal{X}_i)) \not\cong \mathbf{f}^{i,t}_p(H_p(\mathcal{X}_i^m))$ as $t\to\infty$.
\end{thm}
\begin{pf}
Given simplicial complexes are considered, then, for an arbitrary unknown compact topological space $\Top{Z}$, $H_p[(Z,\mathcal{N}_{t}(Z))]=\mathbb{Z}$ for $p=0$ and $H_p[(Z,\mathcal{N}_{t}(Z))]=0$ for $p>0$ as $t\to\infty$. Therefore, if $\mathbf{g}^{0,m}_p(H_p(\mathcal{X}_i))$ is not an isomorphism, as $\Dom{\mathbf{f}^{i,t}_p(H_p(\mathcal{X}_i))} \not\cong \Dom{\mathbf{f}^{i,t}_p(H_p(\mathcal{X}_i^m))}$, i.e., the homology classes of $\Dom{\mathbf{f}^{i,t}_p(H_p(\mathcal{X}_i))}$ which die, or are created after $i$, are different when compared to $\mathbf{f}^{i,t}_p(H_p(\mathcal{X}_i^m))$.
\end{pf}
\begin{corollary}
If $\mathbf{f}^{i,t}_p(H_p(\mathcal{X}_i)) \not\cong \mathbf{f}^{i,t}_p(H_p(\mathcal{X}_i^m))$ there is at least a critical point $k$ between the interval $[i,t)$ in which $\Dom{\mathbf{f}^{k,t}_p(H_p(\mathcal{X}_k)} \cong \Dom{\mathbf{f}^{k,t}_p(H_p(\mathcal{X}_k^m)}$ as $t\to \infty$.
\label{cor:iso-infit}
\end{corollary}
In this sense, considering $t\to\infty$, the $p$-persistence homology $\mathbf{f}^{i,t}_p(H_p(\mathcal{X}_i))$ and $\mathbf{f}^{i,t}_p(H_p(\mathcal{X}_i^m))$ will be equivalent if $\mathbf{g}^{0,m}_p(H_p(\mathcal{X}_i))$ is an isomorphism. Although, even if $\mathbf{g}^{0,m}_p(H_p(\mathcal{X}_i))$ is isomorphic, there is no guarantee that a critical point $k$ between $[i,t)$ will not produce a non-isomorphic $\mathbf{g}^{0,m}_p(H_p(\mathcal{X}_k))$. An adequate representation is then assured when enough points between $[i,j)$ are chosen in order to characterize all possible critical points in the filtration. 

The analysis of the morphism $\mathbf{g}^{0,m}_p$ determines how the topological features change along the levels of a hierarchical clustering model (given by the $p$-persistent homology $\mathbf{f}^{i,t}_p(H_p(\mathcal{X}_i))$) when the HC model is subject to data inclusions. Then, a probability measure $P_\mu$ endowed with a measure $\mu$, somehow associated with $p$-homology groups, is required in order to study the probability that $\mathbf{g}^{0,m}_p(H_p(\mathcal{X}_i))$ is isomorphic whenever $m$ increases. In this sense, the $p$-th Betti-number can be considered such measure as follows: 
\begin{thm}
    The $p$-th Betti-number is a measure over a $\sigma$-algebra given by a collection of all $p$-homology groups.
    \label{theo:expect-betti}
\end{thm}
\begin{pf}
Let $\Top{Z}$ be a topological space, recovering the definition of the $p$-th Betti-number we have $\beta_p := \text{rank}\{H_p[\Top{Z}]\} = \text{rank}\{{\Ima{\partial_{p+1}}/\Ker{\partial_p}}\}$. Considering $p$-dimensional simplicial complex, for a limited natural number $b$, $\Ima{H_p[\Top{Z}]}$ is a free abelian group $\mathbb{Z}^b$ and, therefore, $\text{rank}\{H_p[\Top{Z}]\} = b$. Hence, there is a collection $\mathcal{H} = \{0,\mathbb{Z},\mathbb{Z}^2,\dots\}$, such that $\beta_p$ is a group transformation $\beta_p : (\mathcal{H},\oplus) \rightarrow (\mathbb{N},+)$ where $\oplus$ and $+$ are the direct and the usual sum respectively, and $\beta_p$ is a bijective map. Note that we can construct a $\sigma-$algebra closed under a countable disjoint union $(\phi(\mathcal{H}),\Sigma)$ from the one-by-one mapping $\phi:\mathcal{H} \rightarrow \mathcal{A}$, with $\mathcal{A}=\{A_0,A_1,A_2,\dots | A_i \cap A_j = \varnothing\;\forall\;i \neq j\}$ such that, for a set of indices $E \subset \mathbb{N}$, exists an isomorphism between $\bigoplus_{e \in E} \mathbb{Z}^{h(e)}$ and $\bigsqcup_{e \in E} A_{h(e)}$ with $b = h(e)$ (hence a natural isomorphism for $F:(\mathcal{H},\oplus) \mapsto (\mathcal{A},\sqcup)$). From $\Sigma$, we have the following measure function:
\begin{equation*}
    \mu(A_{h(e)} \in \Sigma) = h(e).
\end{equation*}
which respects
\begin{itemize}
    \item Non-negativity as $h(e) \geq 0$;
    \item Null-empty set with $h(e) = 0$ (induced by $\Ima{\partial_{p+1}C_{p+1}}/\Ker{\partial_p C_p} = \varnothing$);
    \item Countable-additivity as $\mu(\bigsqcup_{e \in E} A_{h(e)}) = \sum_{e \in E} h(e)$.
\end{itemize}
It is worth to mention that these properties are associated one-by-one with the rank of the abelian group $(\mathcal{H},\oplus)$ ($p$-th Betti-number) since:
\begin{itemize}
    \item $\mu(A_{h(e)}) = \text{rank}\{\mathbb{Z}^{h(e)}\}$;
    \item $\mu(\bigsqcup_{e \in E} A_{h(e)}) = \text{rank}\{\bigoplus_{e \in E} \mathbb{Z}^{h(e)}\} = \sum_{e \in E} h(e)$.
\end{itemize}

Therefore, $\beta_p[\Top{X}]$ is a measure endowed of a disjoint set $\sigma$-algebra from which a probability measure $P_{\beta_p}(X)$ can be defined with expected value  $\mathbb{E}_X(\beta_p(X)) = \sum_{k = 1,\dots,\infty} \beta_p(X_k) P_{\beta_p}(X_k)$. 
\end{pf}

Hence, topological features can be measured by $p$-th Betti-numbers, allowing the comparison among topological spaces $\mathcal{X}_i$ built from a Data Clustering or Hierarchical Clustering algorithm. The associated $p$-th Betti-number depends on the inclusion of new i.i.d.\ samples $x'_i$ leading to the stochastic process $\beta_p(\mathcal{X}_i^m|\mathcal{X}_i,x'_1,x'_2,\dots,x'_m)$. In this sense, as $\mathbb{E}\left[\beta_p(\mathcal{X}_i^m)\right] = \beta_p(\mathcal{X}_i)$ is required to guarantee that a DC or HC model is consistent, $\beta_p(\mathcal{X}_i^m)$ should behave as a martingale~\citep{ville1939}, respecting the following properties
\begin{equation*}
    \mathbb{E}\left[\beta_p(\mathcal{X}_i^m)\right] = \beta_p(\mathcal{X}_i^{m-1})\;\text{with}\;\mathbb{E}\left[\beta_p(\mathcal{X}_i)\right] < \infty,
\end{equation*}
which is associated with a difference of martingales in the form
\begin{equation*}
    \mathbb{E}\left[\beta_p(\mathcal{X}_i^m)-\beta_p(\mathcal{X}_i^{m-1})\right] = 0,
\end{equation*}
Consequently, Equation~\ref{eq:mart-diff-seq} calculates the expectation of arising and vanishing $i$-th degree homology classes produced by the inclusion of $m$ new data points:
\begin{equation}
    \mathbb{E}\left[\beta_p(\mathcal{X}_i^m)-\beta_p(\mathcal{X}_i^{m-1})+\dots+\beta_p(\mathcal{X}_i^1)-\beta_p(\mathcal{X}_i)\right] = 0,
    \label{eq:mart-diff-seq}
\end{equation}
implying in
\begin{equation}
    \mathbb{E}\left[\beta_p(\mathcal{X}_i^m)-\beta_p(\mathcal{X}_i)\right] = 0.
    \label{eq:mart-mexp}
\end{equation}

As foretold, $\beta_p(\mathcal{X}_i^m)$ is considered to behave as a martingale, hence Azuma's Inequality~\citep{azuma1967} (Inequality~\ref{eq:mart-convergence} in~\ref{app:proofs}) allows the study of the convergence of the following generalization measure: 
\begin{equation}
|\beta_p(\mathcal{X}_i^m)-\beta_p(\mathcal{X}_i)|,
\label{eq:mult-criteria}
\end{equation}
and, therefore, of $\mathbb{E}\left[\beta_p(\mathcal{X}_i^m)-\beta_p(\mathcal{X}_i)\right]$. Employing such generalization we formulate the DC $p$-homology convergence Lemma (the proof is in~\ref{app:proofs}):
\begin{lemma}[DC $p$-homology convergence]
    Let $X$ and $X^m$ be two datasets with i.i.d. samples such that $X^m = X \cup \{x'_1,\dots,x'_m\}$, $\mathcal{X}_i$ be the neighborhood topological space built up from a points cloud, and $\beta_p(\cdot)$ be the $p$-th Betti-number calculated upon the neighborhood topologies. The probability that the average absolute difference $\sum_{q=1}^{m} |\beta_p(\mathcal{X}_i^q) - \beta_p(\mathcal{X}_i)|$ is bounded by $\epsilon$ decays exponentially with respect to
    \begin{equation}
    P\left(\sum_{q=1}^{m} |\beta_p(\mathcal{X}_i^q) - \beta_p(\mathcal{X}_i)| > m\epsilon\right) \leq 2\exp\left(\frac{-m^2\epsilon^2}{2\sum_{q=1}^{m} c_{q,p}^2}\right),
    \label{eq:crit-prob-convergence}
    \end{equation}
    given $c_{q,p} = |\beta_p(\mathcal{X}_i^q)-\beta_p(\mathcal{X}_i)|$ is bounded.
    \label{lemma:DC-prob-conv}
\end{lemma}
Then, whenever the difference between the number of $p$-homology classes of $\mathcal{X}_i^{q}$ and $\mathcal{X}_i$ increases asymptotically faster than $m$, there is no guarantee that, in average, $\beta_p(\mathcal{X}_i^q)$ approximates $\beta_p(\mathcal{X}_i)$. Conversely, if $|\beta_p(\mathcal{X}_i^q)-\beta_p(\mathcal{X}_i)|$ is asymptotically smaller than $m$, $|\beta_p(\mathcal{X}_i^m)-\beta_p(\mathcal{X}_i)| \to 0$, and, therefore, $\mathbf{g}_{0,m}(\mathcal{X}_i)$ is likely to produce an isomorphism. 

\begin{rmk}
For instance, considering the case of $H_0$, if each new sample increases the number of connected components by one, i.e., if an over-refinement occurs, the terms $c_{q,p}$ grow as follows:
\begin{equation*}
    \sum_{q=1}^{m} c_{q,0}^2 = \sum_{q=1}^{m} q^2 = \frac{1}{6} m (m + 1) (2m + 1),
\end{equation*}
consequently, in this case, Inequality~\ref{eq:crit-prob-convergence} diverges as follows:
\begin{equation*}
    P\left(\sum_{q=1}^{m} |\beta_p(\mathcal{X}_i^q) - \beta_p(\mathcal{X}_i)| > m\epsilon\right) \leq 2\exp\left(\mathcal{O}(m^{-1} )\right),
\end{equation*}
hence, there is no guarantee that $\mathbf{g}_{0,m}(\mathcal{X}_i)$ is isomorphic. 
\end{rmk}

\begin{corollary}
An over-refinement occurs whenever
\begin{equation*}
    \mathcal{\mathbf{\Omega}} \left(c_{q,p}\right) = m,
\end{equation*}
and convergence of the right-hand term in Inequality~\ref{eq:crit-prob-convergence} occurs whenever
\begin{equation*}
    \mathcal{O} \left(c_{q,p}\right) < \sqrt{m}.
\end{equation*}
\label{cor:DC-bound}
\end{corollary}
Therefore, whenever $\mathcal{O} \left(c_{q,p}\right) < \sqrt m$, $\mathbf{g}_{0,m}(\mathcal{X}_i)$ is likely to produce an isomorphism. In this sense, a maximum value $\overline{c}_{p,q}$ bounded as described in Corollary~\ref{cor:DC-bound} also guarantees consistency.

%
\begin{corollary}
    Let $\overline{c}_{q,p} = \mathrm{max}_{q=1,\dots,m}(c_{q,p})$ and $m$ be the number of new samples added to a dataset $X$, 
\begin{equation}\label{eq:max_cp}
    \overline{c}_p < \epsilon \sqrt{m/2\ln{2}}.
\end{equation}
\label{cor:DC-max_c}
\end{corollary}

The result of Lemma~\ref{lemma:DC-prob-conv} can, therefore, be extended for the HC problem. Remember that every simplicial complex of $\mathcal{X}_i^r$ and $\mathcal{X}_i^q$ in the filtrations, with $r,q=1,\dots,m$, must be associated with the same pre-image in the tame function $f$. In this, the convergence for HC models in terms of $p$-th Betti-numbers is given respecting the following theorem
\begin{thm}[HC $p$-homology convergence]
Let $\Ima{f}^{-1}$ be the pre-image of the tame function $f$ associated with the filtrations $\mathcal{X}_i^q\subseteq \dots \subseteq \mathcal{X}_j^q$, and a bijective function $h:\Ima{f}^{-1} \rightarrow \mathbb{N}$ which maps the points of $\Ima{f}^{-1}$ onto the index of such filtrations. Then, the DC $p$-homology convergence for each one of those points is given by
\begin{equation*}
    P\left(\sup_{t \in \Ima{f}^{-1}} \sum_{q=1}^{m} |\beta_p(\mathcal{X}_{h(t)}^q) - \beta_p(\mathcal{X}_{h(t)})| > m\epsilon\right) \rightarrow 0\;\text{as}\;m \rightarrow \infty
\end{equation*}
with
\begin{equation*}
 P\left(\sup_{t \in \Ima{f}^{-1}} \sum_{q=1}^{m} |\beta_p(\mathcal{X}_{h(t)}^q) - \beta_p(\mathcal{X}_{h(t)})| > m\epsilon\right) \leq 2M \exp\left(\frac{-m\epsilon^2}{2\tilde{c}_p^2}\right),
\end{equation*}
where $M$ is the cardinality of the pre-image $\Ima{f}^{-1}$ of a tame function $f$, and $\tilde{c}_p = \max_{i \in \mathbb R} \overline{c}_{p,i}$.
\label{theo:HC-conv}
\end{thm}
\begin{corollary}
    Considering Theorem~\ref{theo:HC-conv}, let us define $\overline{\Delta\beta}_{p,t}=\sum_{q=1}^{m} |\beta_p(\mathcal{X}_{h(t)}^q) - \beta_p(\mathcal{X}_{h(t)})|$ and take $\delta = P\left(\sup_{t \in \Ima{f}^{-1}} \overline{\Delta\beta}_{p,t} \leq m\epsilon\right)$, then the HC $p$-homology convergence is guaranteed whenever
    \begin{equation}
    \frac{\tilde{c}_p^2 \ln{M}}{m} \rightarrow 0 \text{\;\;as\;\;} m\to\infty.
    \label{eq:conv-criteria}
    \end{equation}
\end{corollary}
\begin{corollary}
If $\tilde{c}_p$ is constant, convergence of Equation~\ref{eq:conv-criteria} is equivalent to
\begin{equation*}
\frac{\ln{M}}{m} \rightarrow 0 \text{\;\;as\;\;} m\to\infty.
\end{equation*}
\end{corollary}
\begin{rmk}
The parameter $\tilde{c}_p$ can be stated as constant, for example, such as in partitional clustering as K-means~\citep{Steinhaus56,Macqueen67}, K-medoids~\citep{kaufmanl1987} and Self-Organizing Maps~\citep{Kohonen1982} which relies on optimization procedures, typically based on some distance from data points and pre-determined centroids. Considering an i.i.d. data distribution and enough samples, those centroids will never sufficiently vary as new data are acquired, hence producing the same number of connected components. Although, as the number of centroids increases w.r.t. the new samples acquired, it is likely that those algorithms identify every point as a single cluster, and HC $0$-homology consistency is not guaranteed. 
\end{rmk}

Considering the $H_0$ homology group, the generalization is guaranteed for every critical point greater than the first point that ensures the DC $0$-homology consistency, as proved in the following lemma:
\begin{lemma}
Assuming the $H_0$ homology group, we have that, for every critical point $i < j$, which implies in $\beta_0(\mathcal{X}_i^q) > \beta_0(\mathcal{X}_j^q)$, if $\mathcal{X}_i$ is DC $0$-homology consistent then $\mathcal{X}_j$ is also DC $0$-homology consistent.
\label{lemma:seq-consistency}
\end{lemma}
\begin{pf}
    As $\mathcal{X}_i^q$ is DC $0$-homology consistent, following Lemma~\ref{lemma:DC-prob-conv}, we have that $\textnormal{max}_{q=1,\dots,m}(|\beta_p(\mathcal{X}_i^q)-\beta_p(\mathcal{X}_i)|)$ must have to be bounded and, therefore, $\beta_p(\mathcal{X}_i^q)$ and $\beta_p(\mathcal{X}_i)$ are also bounded. In this sense, as $\beta_p(\mathcal{X}_i^q),\beta_p(\mathcal{X}_i) \geq 0$, $\beta_p(\mathcal{X}_j^q) \leq \beta_p(\mathcal{X}_i^q)$ and $\beta_p(\mathcal{X}_j) \leq \beta_p(\mathcal{X}_i)$, then $\textnormal{max}_{q=1,\dots,m}(|\beta_p(\mathcal{X}_j^q)-\beta_p(\mathcal{X}_j)|) < \infty$ implying that $\mathcal{X}_j$ is stable, hence DC $0$-homology consistent.
\end{pf}
Following these consistency results we conclude that~\citet{kleinberg02}'s richness axiom must be relaxed as it considers partitions which does not guarantee $p$-homology consistency. Section~\ref{sec:carlsson} will discuss how our results are related with~\citet{carlsson10}'s consistency for their adapted single-linkage method.

\section{An ultrametric perspective of Coarse-Refinement dilemma}\label{sec:carlsson}

In~\citep{carlsson10}, the hirarchical clusters of the agglomerative algorithm single-linkage are described as dendrograms which are equivalent to ultrametric spaces $(X,u_d)$. Those ultrametric spaces were employed in order to support~\citet{carlsson10} study of the convergence and the stability for their proposed version of the single-linkage algorithm. In this sense, a HC model of single-linkage approximates the unknown underlying ultrametric space following the Gramov-Hausdorff distance~\citep{gromov1981} as the number of samples $m$ increases. Hence, $m$ increases, the HC model becomes isometric to the underlying ultrametric space. 

More precisely, in~\citep{carlsson10}, a hierarchical clustering algorithm produces a compact metric space $(A,d_A)$ from which $A$ is a finite collection of disjoint compact subsets $\{U^{(\alpha)}\}_{\alpha\in A}$ of a compact metric space $(Z,d_Z)$. The distance function of such metric space $(A,d_A)$ is defined by a linkage function in form $d_A:=\mathcal{L}(W_A)$, from which $W_A:A\times A \to \mathbb{R}^+$ is given by $(\alpha,\alpha') \mapsto \text{min}_{z\in U^{(\alpha)},z'\in U^{(\alpha')}}$. The minimal separation of $(A,d_A)$ is then defined as
\begin{equation*}
    \text{sep}(A,d_A) := \text{min}_{\alpha,\alpha' \in A,\alpha\neq \alpha'} W_A,
\end{equation*}
which calculates the minimum distance between $U^{(\alpha)}$ and $U^{(\alpha')}$.

In order to formulate the probabilistic convergence for single-linkage,~\citet{carlsson10} consider a hierarchical clustering model on metric measure spaces. In this sense, the measurable space $(X,d_X,\mu_X)$, endowed with a metric space $(X,d_X)$ and a Borel probability measure $\mu_X$ on $X$ with compact support $\text{supp}(\mu_X)$, are defined as an mm-space (measure metric space)~\citep{gromov99}. The authors also define a function $f_X:\mathbb{R}^{+} \to \mathbb{R}^+$ as $r \mapsto \text{min}_{x\in\text{supp}(X)} \mu_X(B_X(x,r))$, which is non-decreasing and $f(X) > 0$ for all $r > 0$. In this sense, let $F_X:\mathbb{N}\times \mathbb{R}^+ \to \mathbb{R}^+$ be a function defined by $(n,\delta) \mapsto \frac{e^{-m f_X(\delta/4)}}{f_X(\delta/4)}$. Note that~\citet{carlsson10} assume $\delta_0$ to be fixed and, therefore, $F_X(\cdot,\delta_0)$ to be bounded and decreasing, in order to prove the following theorem
\begin{thm}
    Let $(Z,d_Z,\mu_Z)$ be an mm-space and $\text{supp}=\bigcup_{\alpha \in A} U^{(\alpha)}$ for a finite index set $A$ and $\mathbf{U} = \{U^{(\alpha)}\}$ a collection of disjoint, compact, path-connected subsets of $Z$. Let $(A,d_A)$ be the metric space arising from $\mathbf{U}$ and let $\delta_A:=\text{sep}(A,d_A)/2$.
    For each $n \in \mathbb{N}$, let $X=\{x_1,x_2,\dots,x_n\}$ be a collection of $n$ independent random variables with distribution $\mu_Z$ and let $d_X$ be the restriction of $d_Z$ to $X \times X$. Then, for $\zeta \geq 0$ and $n \in \mathbb{N}$,
    \begin{equation}
        P_{\mu_Z}\left(d_{GH}\big(\mathfrak{L}(X,d_X),\mathfrak{L}(A,d_A)\big) > \zeta\right) \leq F_Z(n,\text{min}(\zeta,\delta_A/2)),
        \label{eq:carlson-conv}
    \end{equation}
    with $\mathfrak{L}$ being the ultrametric space produced by~\citet{carlsson10}'s single linkage algorithm.
    \label{theo:theorem-carlsson}
\end{thm}

Although it is assumed that $\delta_0$ is fixed and, therefore, Inequality~\ref{eq:carlson-conv} converges. Although, as $B_X(x,r)$ reduces, $\mu_X(B_X(x,r))$ tends to be uniform as $B_X(x,r)$ induces the trivial topology and, therefore, $f_X(\epsilon)=1/m$ with an infinitesimal epsilon. 
We prove that, given $\beta_0(\mathcal{X}_\delta)$ the number of connected components in the clustering associated with the radius $\delta$, if $\beta_0(\mathcal{X}_\delta) \ln{\beta_0(\mathcal{X}_\delta)}$ reduces w.r.t. $m$, so that~\citet{carlsson10}'s consistency is guaranteed, i.e., if the $0$-dimensional homology groups increase, thus Inequality~\ref{eq:carlson-conv} does not converge to zero.

\begin{thm}
    If $\frac{\beta_0(\mathcal{X}_{\delta/4}) \ln \beta_0(\mathcal{X}_{\delta/4})}{m}$ diverges as $m\to \infty$ ~\citet{carlsson10}'s consistency is not guaranteed and if $\frac{\beta_0(\mathcal{X}_{\delta/4})^2}{m}$ converges then~\citet{carlsson10}'s consistency holds.
\end{thm}

\begin{pf}
In the first step of the proof for Theorem~\ref{theo:theorem-carlsson},~\citet{carlsson10} study the probabilistic convergence, in terms of the Hausdorff distance, between a restricted mm-space $\mathcal{S}_{mm}(X) = (X,d_X,\mu_X)$ and the support of $\mu_X$ as
\begin{equation*}
    P_{\mu_Z}\left(d^X_H(X,\text{supp}(\mu_X)>\delta)\right) \leq F_X(n,\delta).
\end{equation*}
On further formulation, they find that
\begin{equation*}
     P_{\mu_Z}\left(d^X_H(X,\text{supp}(\mu_X))>\delta\right) \leq N e^{-m f_X(\delta/4)},
     \label{eq:part-conv-carlson}
\end{equation*}
with $N$ equals to the cardinality of the maximal $\delta/4$-packing of $\text{supp}(\mu_X)$. A $\delta$-packing is defined as
\begin{defn}
     Let $(\mathfrak{M},d_\mathfrak{M})$ be a metric space and $\mathcal{M} \subset \mathfrak{M}$. $\{\mathcal{M}_1,\dots,\mathcal{M}_n\}$ is an $\delta$-packing of $\mathcal{M}$ if $\bigcap_{i=1}^n B(\mathcal{M}_i,\delta)=\varnothing$.
\end{defn}

Therefore, the topologial space induced by the space $(\mathfrak{M},d_\mathfrak{M})$ is a Hausdorff space and its cardinality is given by its connected components $\beta_0(\mathcal{X}_{\delta/4})$ implying that~\citet{carlsson10}'s consistency implicitly considers the number of $0$-dimensional homology classes in its formulation. Then, let $\mathfrak{L}(\cdot,\cdot)$ be defined as the single-linkage function, whenever the induced neighborhood topology is over-refined, there is no guarantee that Inequality~\ref{eq:carlson-conv} will converge since
\begin{equation*}
    P_{\mu_Z}\left(d_{GH}\big(\mathfrak{L}(X,d_X),\mathfrak{L}(A,d_A)\big) > \zeta\right) \leq P_{\mu_Z}\left(d^X_H(X,\text{supp}(\mu_X))>\zeta)\right),
\end{equation*}
as proved in~\citep{carlsson10}.

More precisely, as $1/f_X(\delta/4)$ is a superior bound for $N$~\citep{carlsson10}, then
\begin{equation*}
    \frac{\beta_0(\mathcal{X}_{\delta/4}) \ln \beta_0(\mathcal{X}_{\delta/4})}{m} \leq \frac{\ln N}{m f_X(\delta/4)},
\end{equation*}
so divergence of Inequality~\ref{eq:part-conv-carlson} occurs whenever
\begin{equation*}
    \frac{\beta_0(\mathcal{X}_{\delta/4}) \ln \beta_0(\mathcal{X}_{\delta/4})}{m}\to \infty\;\textnormal{as}\;m\to \infty.
\end{equation*}
Hence, if the number of connected components ($0$-dimensional homology classes) increases with the number of data points $m$, there is no convergence guarantee for~\citet{carlsson10}'s consistency, i.e., if the clustering model is not DC $0$-homology consistent (Definition~\ref{def:dc-p-hom-cons}), then~\citet{carlsson10}'s consistency is not guaranteed. 
In order to prove the upper bound for~\citet{carlsson10}'s consistency, we apply Taylor's Theorem in $\ln{N}$ resulting in
\begin{equation*}
    \frac{\ln N}{m f_X(\delta/4)} \leq \frac{N}{m f_X(\delta/4)} - \frac{1}{m f_X(\delta/4)},
\end{equation*}
then, taking
\begin{equation*}
    \frac{\beta_0(\mathcal{X}_{\delta/4})^2}{m} \leq \frac{N}{m f_X(\delta/4)},
\end{equation*}
implies
\begin{equation*}
    \frac{\ln N}{m f_X(\delta/4)} \leq \frac{\beta_0(\mathcal{X}_{\delta/4})^2}{m} - \frac{1}{m f_X(\delta/4)} \leq \frac{\beta_0(\mathcal{X}_{\delta/4})^2 - \beta_0(\mathcal{X}_{\delta/4})}{m}.
\end{equation*}
Therefore, if DC $0$-homology convergence ($\frac{\beta_0(\mathcal{X}_{\delta/4})^2}{m} \leq \frac{\overline{c}_{p}^2}{m} \to 0\;\text{as}\;m\to \infty$ in Inequality~\ref{eq:max_cp}) is guaranteed then~\citet{carlsson10}'s consistency holds and, if~\citet{carlsson10}'s consistency diverges, DC $0$-homology convergence does not holds. 
Following the Lemma~\ref{lemma:seq-consistency}, as the considerd space is the ultrametric one, those consistencies are ensured for all radius $r$ greater than $\delta/4$ determined for the single-linkage algorithm. In addition, as~\citet{carlsson10}'s consistency does not hold in over-refined metric spaces, also the $H_0$ homology group of its induced topology will not be DC $0$-homology consistent. 
\end{pf}

Section~\ref{sec:exps} provides experimental results and analysis considering the generalization measure to illustrate how homology classes diverge when new samples are included from the same topological space and probability distribution. 

\section{Experiments and results}\label{sec:exps}

The experiments discussed in this section rely on five different datasets: a torus (Toy experiment), the crescent moon dataset (Experiment I), samples collected from a Lorenz attractor (Experiment II), from a R\"{o}ssler attractor (Experiment III) and, finally, from the Mackey Glass attractor (Experiment IV). All these experiments were conducted to study the validity of our proposed generalization measure (defined in Equation~\ref{eq:mult-criteria}).

\subsection{Experimental setup}\label{subsec:expsmethod}

In order to perform all experiments, homology simplicial complexes based on open balls, such as Vietori-Rips~\citep{Hausmann95} and Lazy Witness~\citep{DeSilva:2004}, were used to compute the persistences associated to the derived homology classes. Note that the Vietori-Rips method requires extensive computational processing, therefore the Lazy Witness complex was adopted in large datasets in order to reduce computational complexity. In addition, landmarks were chosen from the original dataset and maintained in the analysis of the perturbed one in order to preserve the inclusions in the bifiltration. It is worth to mention that the experiments in which Lazy Witness was employed are sensitive to point density~\citep{DeSilva:2004} when landmarks are chosen randomly. Then, counterintuitively, the number of connected components for the minimum radius employed decreased as the number of included samples $m$ increased because new witnesses were identified and, therefore, prior connected components were merged by such new instances. Also, this affects the asymptotical convergence of the barcode plots~\citep{carlsson09}: whenever new samples are acquired, a connected component is more likely to be merged and, therefore, the barcode plot converges to zero faster in the presence of perturbed data.

Considering the experimental setup, complexes were created based on the spatial configuration of open balls with radius $r$ centered at data points. Complexes produced throughout some range $[0,r)$ define a filtration along which we verify the birth and death of homology classes in order to represent the persistence intervals. From such interval, for a set of radius $r_0 < r_1 < \dots < r_i < \dots < r_\text{max}$, $\beta_p(\mathcal{X}_{r_i})$ and $\beta_p(\mathcal{X}_{r_i}^m)$ were computed in order to calculate $|\beta_p(\mathcal{X}_{r_i}) - \beta_p(\mathcal{X}_{r_i}^m)|$. Assuming $W_{r_i} = |\beta_p(\mathcal{X}_{r_i}) - \beta_p(\mathcal{X}_{r_i}^m)|$, the space $W_{r_i} \times r_i$ provides how many homology classes are preserved for each radius $r_i$, allowing the study of $W_{r_i}$ along the filtration. In addition, in order to analyze the convergence, we also partitioned the perturbed set $X^m$ into $X_U = \{X^{m_1},X^{m_2},\dots,X^{m}\}$ such that $m_1 < m_2 < \dots < m \implies X^{m_1} \subset X^{m_2} \subset \dots \subset X^{m}$, creating a space $W_{r_i} \times r_i \times M$ that represents the generalization measure values along different sampling and radii. More precisely, $X^m$ was set up with $10,000$ samples and $m_i$ assumes values between $5,100$ and $10,000$ with a unity step equals to $200$. From such results, we plotted a heatmap in order to study the convergence of $W_{r_i}$ along data inclusions.

The following subsections detail the experiments and discuss their results.

\subsection{Toy experiment - Bidimensional torus}

Let a topological space $\Top{X}$ compose a torus with internal and external radii of, respectively, $0.5$ and $1$ units, given $400$ samples. It was generated using the following function
\begin{equation}
    f(R,r,\theta,\phi) = \left((R + r\cos{\theta}) \cos{\phi},(R + r\cos{\theta})\sin{\phi}\right),
    \label{eq:tor}
\end{equation}
in which $\phi$ and $\theta$ assume uniformly sampled values from the interval $[0,2\pi]$, producing the dataset illustrated in Figure~\ref{fig:data-torus} of~\ref{app:datasets}. The perturbed data $X^m$ was produced by the inclusion of $100$ new observations into $X$, maintaining the same parametric setup and, consequently, the data distribution, as discussed before.

\begin{figure}[htb!]
\centering
\includegraphics[scale=0.64]{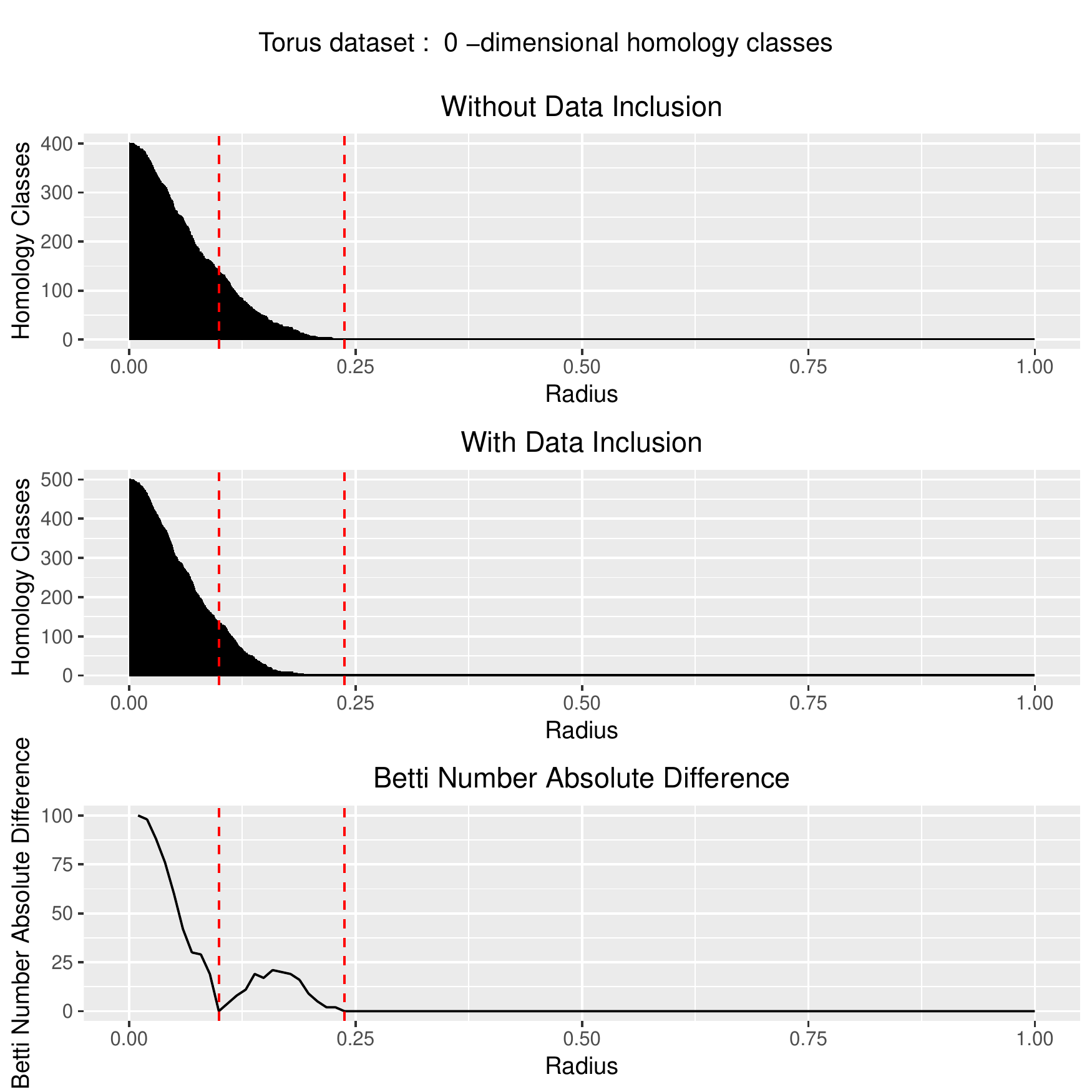}
\caption{Graphs produced from $0$-dimensional homology classes which correspond to, from top to bottom: i) Barcode plots generated over the torus experiment with $400$ samples; ii) Barcode plots generated over the perturbed dataset with $500$ samples; and, finally, iii) the values for the generalization measure $|\beta_0(\mathcal{X}_{r_i}) - \beta_0(\mathcal{X}_{r_i}^m)|$. The red-dashed lines mark the initial value of the intervals which ensure $|\beta_0(\mathcal{X}_{r_i}) - \beta_0(\mathcal{X}_{r_i}^m)| = 0$.}
\label{fig:barcode-torus-0}
\end{figure}

We then applied the Vietori-Rips~\citep{Hausmann95} filtration (available in the TDA Package from the R Project for Statistical Computing~\footnote{\url{https://cran.r-project.org/web/packages/TDA}}) on each dataset, $X$ and $X^m$, setting the radius interval as $[0,1)$ in order to produce barcode plots with zero and one dimensions, as illustrated in Figure~\ref{fig:barcode-torus-0}. Then we computed space $W_{r_i} \times r_i$, as aforementioned in Section~\ref{subsec:expsmethod}, with $r_i$ assuming the values $\{0.01,0.02,\dots,0.99,1\}$ (steps of $0.01$). This resulted in the bottom graph of Figure~\ref{fig:barcode-torus-0}.

\begin{figure}[htb!]
\centering
\includegraphics[scale=0.64]{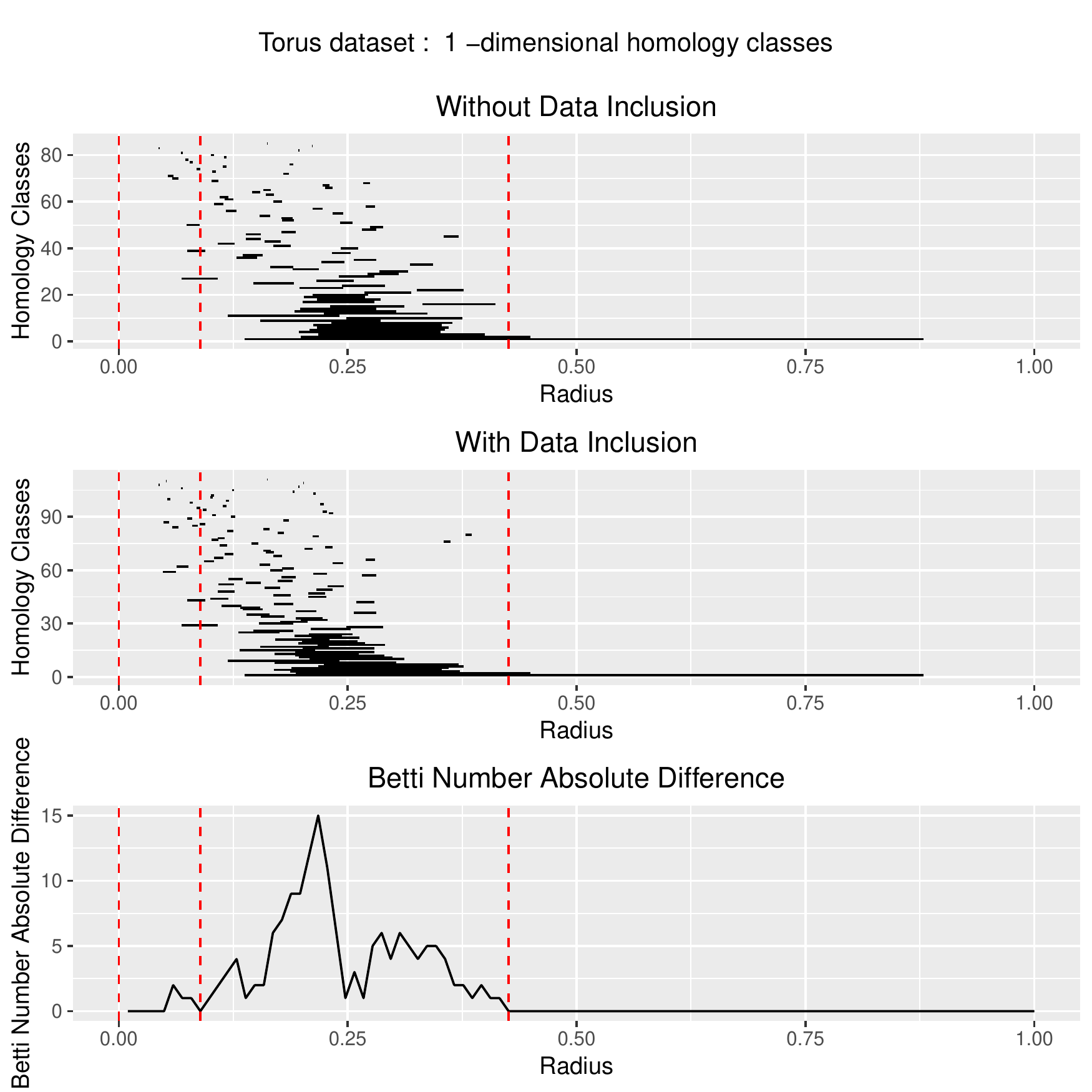}
\caption{Graphs produced from one-dimensional homology classes which correspond to, from top to bottom: i) Barcode plots generated over the torus experiment with $400$ samples; ii) Barcode plots generated over the perturbed dataset with $500$ samples; and, finally, iii) the values for the generalization measure $|\beta_1(\mathcal{X}_{r_i}) - \beta_1(\mathcal{X}_{r_i}^m)|$. The red-dashed lines mark the initial value of the intervals which ensure $|\beta_1(\mathcal{X}_{r_i}) - \beta_1(\mathcal{X}_{r_i}^m)| = 0$.}
\label{fig:barcode-torus-1}
\end{figure}

As illustrated in Figure~\ref{fig:barcode-torus-0}, the minimal values are achieved when the radius $r_i$ is equal to $0.01$ and when it is contained in interval $[0.2376,1)$. This result indicates that such radii values may produce the same number of $0$-dimensional homology classes, i.e., connected components, for both datasets. In addition, when one-dimensional homology classes are studied, the first minimal values, in the radius range $[0,0.0891)$, should be disregarded because, as illustrated in Figure~\ref{fig:barcode-torus-1}, there are no homology classes of degree one in such interval. Nonetheless, the radius interval $[0.4257,1)$ presents the same number of one-dimensional homology classes for $X$ and $X^m$. As illustrated in Figure~\ref{fig:barcode-torus-1}, the most relevant homology class vanishes when the radius approximates $0.88$ as the one-dimensional hole of the torus is filled out.

\subsection{Experiment I - Crescent moon dataset}

The Crescent Moon dataset, available with the RSSL package from the R Project for Statistical Computing~\footnote{\url{https://cran.r-project.org/web/packages/RSSL}}, was produced employing the function {\it generateCrescentMoon} using parameters $n=5,000$ ($10,000$ samples were produced, given $n$ is associated with the number of points per class and it considers the binary problem), $d=2$ and $\sigma=0.5$, then producing the observations illustrated in Figure~\ref{fig:crescmoon} of~\ref{app:datasets}.

At a next step, we applied the Lazy Witness method~\citep{DeSilva:2004} to define a filtration with radius interval set as $[0,3)$ in order to calculate the simplicial complexes with degree one, adopting $200$ landmarks. Then, we computed the space $W_{r_i} \times r_i$, as discussed in Section~\ref{subsec:expsmethod}, assuming $10,000$ values for $r_i \in [0,3)$. As illustrated in Figure~\ref{fig:moon-barcodes-0} of~\ref{app:barcodes}, for $0$-dimensional complexes, the radius interval $[0,0.84)$ depicts differences between $\beta_0(\mathcal{X}_{r_i})$ and $\beta_0(\mathcal{X}_{r_i}^m)$, and the generalization measure does not converge to zero. As illustrated also in Figure~\ref{fig:moon-barcodes-0} (\ref{app:barcodes}), radius values in interval $[x,y)$ allowed to identify only one cluster in the dataset. This is caused by the influence of noise located among clusters, which may link one connected component to another. Such issues motivate the employment of density-based clustering techniques such as DBSCAN~\citep{Ester96}.

\begin{figure}[htb!]
\centering
\includegraphics[scale=0.64]{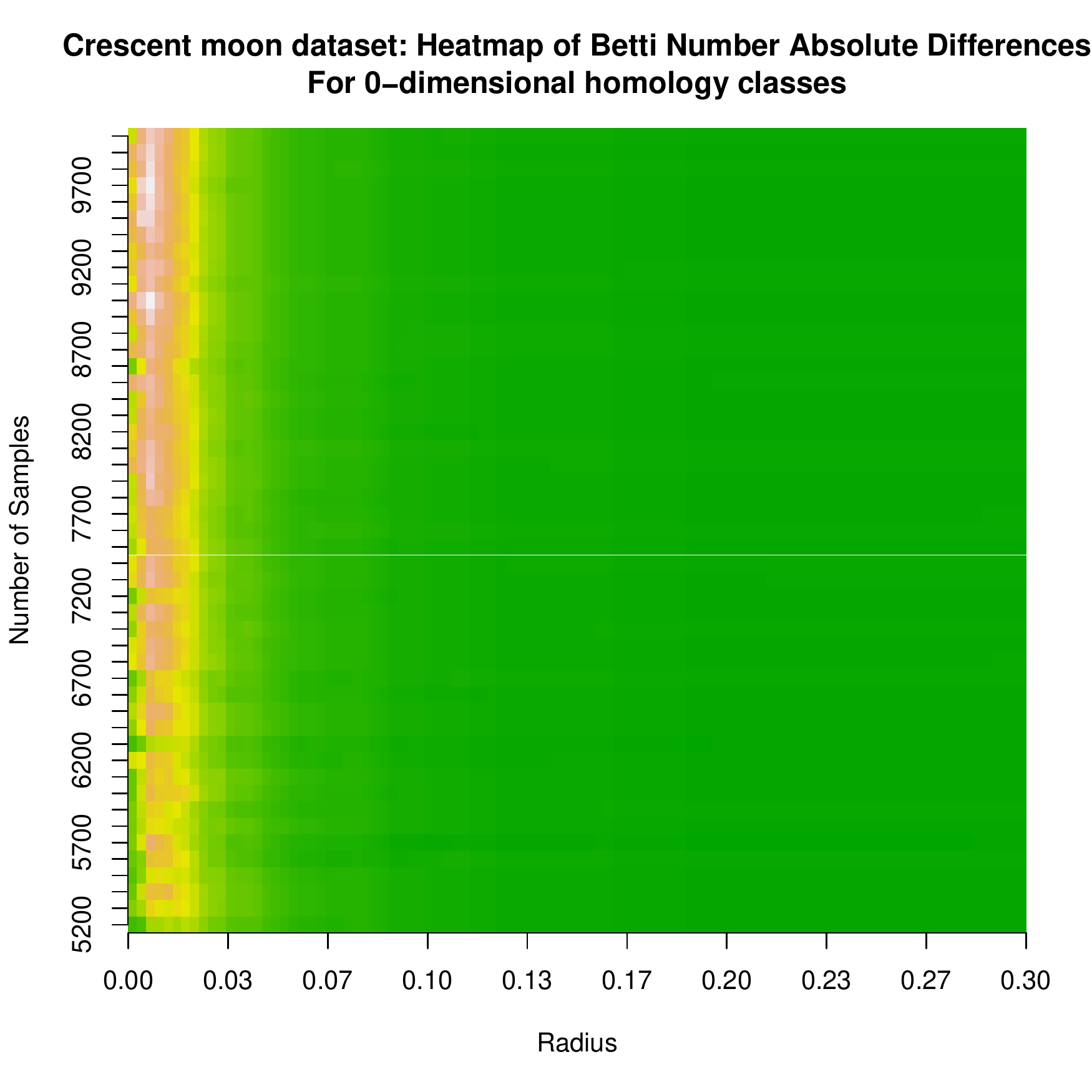}
\caption{The heatmap generated from the Crescent Moon dataset formed by values of the generalization measure $|\beta_0(\mathcal{X}_{r_i}) - \beta_0(\mathcal{X}_{r_i}^m)|$ along the insertion of samples and the increase in radius in the filtration.}
\label{fig:banana-heatmap-0}
\end{figure}

In addition, Figure~\ref{fig:banana-heatmap-0} illustrates how $W_{r_i}$ increases along an approximate radius interval of $[0,0.03)$, and how its values stay constant on the complementary interval. In this sense, each small radius does not guarantee isomorphisms among homology classes as new data is included in this dataset.

\subsection{Experiment II - Lorenz attractor}

The Lorenz attractor dataset was produced using the Lorenz system implemented in the package nonlinearTseries from the R Project for Statistical Computing~\footnote{Package nonlinearTseries is available at \url{https://cran.r-project.org/web/packages/nonlinearTseries}}. The parameters employed to generate the attractor were $\sigma=10$, $\beta=8/3$, $\rho=28$ with initial conditions $x_0=-13,y_0=-14$, and $z=47$. The parameter time was set as a sequence from $0$ to $50$ given steps of $0.005$ units, producing the attractor shape illustrated in Figure~\ref{fig:lorenz} (\ref{app:datasets}).

Lazy Witness, employing $200$ landmarks, was then applied to define a filtration using the radius interval $[0,7)$ to form simplicial complexes with degree equal at most to one. The space $W_{r_i} \times r_i$ was computed assuming $10,001$ values for $r_i \in [0,7)$. As illustrated in Figure~\ref{fig:lorenz-barcodes-0} (\ref{app:barcodes}), interval the $[3.829,7)$ respects the proposed generalization measure for $0$-dimensional homology classes, guaranteeing the production of a single connected component as expected.

\begin{figure}[htb!]
\centering
\includegraphics[scale=0.64]{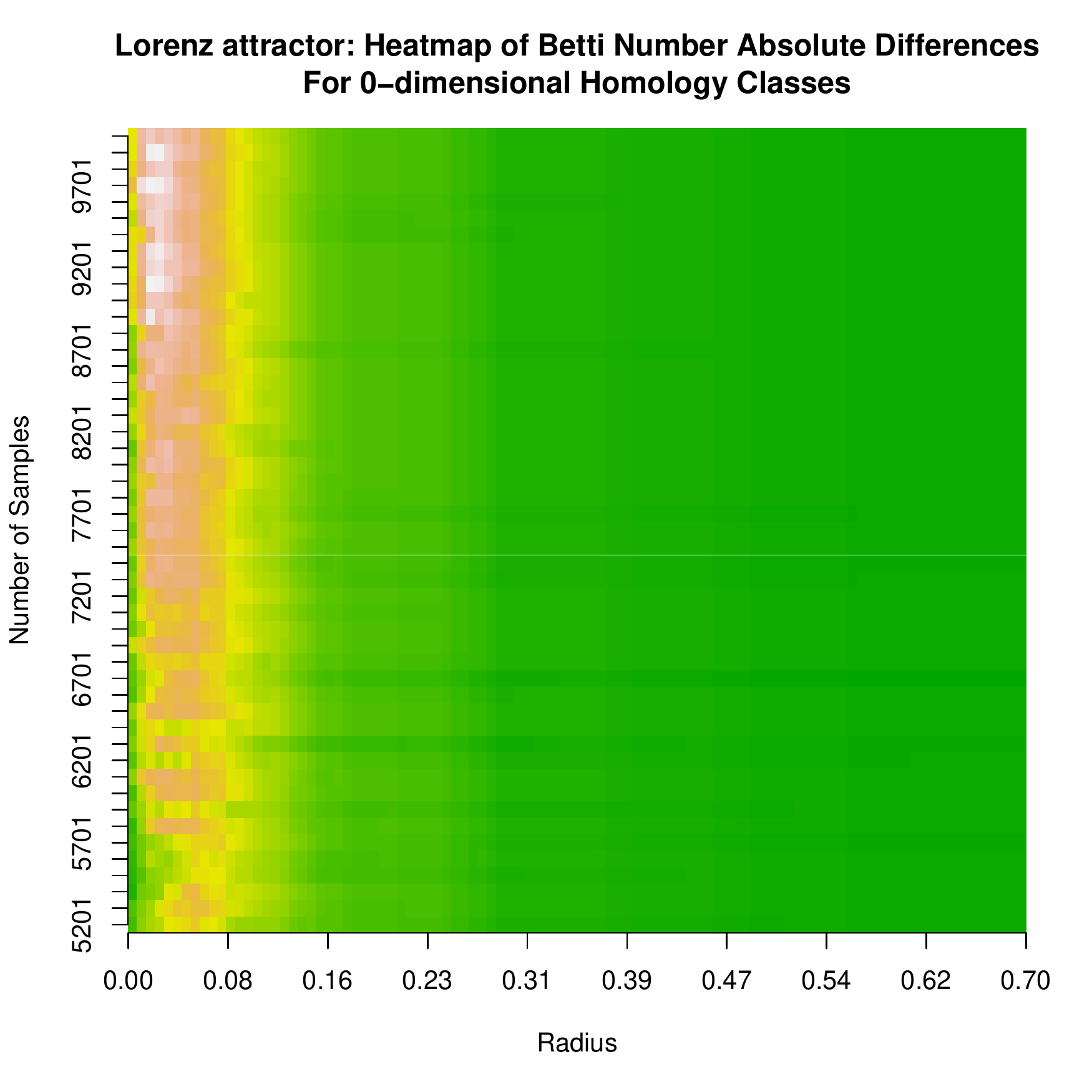}
\caption{The heatmap generated from the Lorenz attractor formed by values of the generalization measure $|\beta_0(\mathcal{X}_{r_i}) - \beta_0(\mathcal{X}_{r_i}^m)|$ along the insertion of samples and the increase of radius in the filtration.}
\label{fig:lorenz-heatmap-0}
\end{figure}

The heatmap, shown in Figure~\ref{fig:lorenz-heatmap-0}, is used to analyze the convergence of $W_{r_i}$, given the inclusion of new data as the radius changes. By analyzing this heatmap, we verify that the first radius interval presents the same increasing behavior reported in previous experiments, providing additional evidences that our Coarse-Refinement dilemma holds analogously to as the Bias-Variance dilemma~\citep{Geman92,luxburg09}.

Considering the one-dimensional homology classes, the set of intervals respecting $|\beta_1(\mathcal{X}_{r_i})-\beta_1(\mathcal{X}_{r_i}^m)| = 0$ are $[1.2747,1.4280)$, $[1.6037,3.9333)$, $[4.4590,\break 5.5643)$, and $[5.6420,7)$. As illustrated in Figure~\ref{fig:lorenz-barcodes-1} (\ref{app:barcodes}), the first two intervals are enough to represent the two attractor cycles, as expected. Interval $[3.9333,4.4590)$ also supports the identification of two cycles if one considers the first $5,101$ dataset observations; this changes to a single cycle when $4,900$ extra observations are added up. In sequence, any radius in $[4.4590, 5.5643)$ produces a single $1$-dimensional homology class on both datasets (with $5,101$ and $10,001$ observations each). On the other hand, using the interval $[5.5643, 5.6420)$, a single $1$-dimensional homology class will be produced for the dataset with $5,101$ samples, and no one for the other dataset given the same dimensionality. Finally, the last interval will not generate any $1$-dimensional homology class for those two datasets.

\begin{figure}[htb!]
\centering
\includegraphics[scale=0.64]{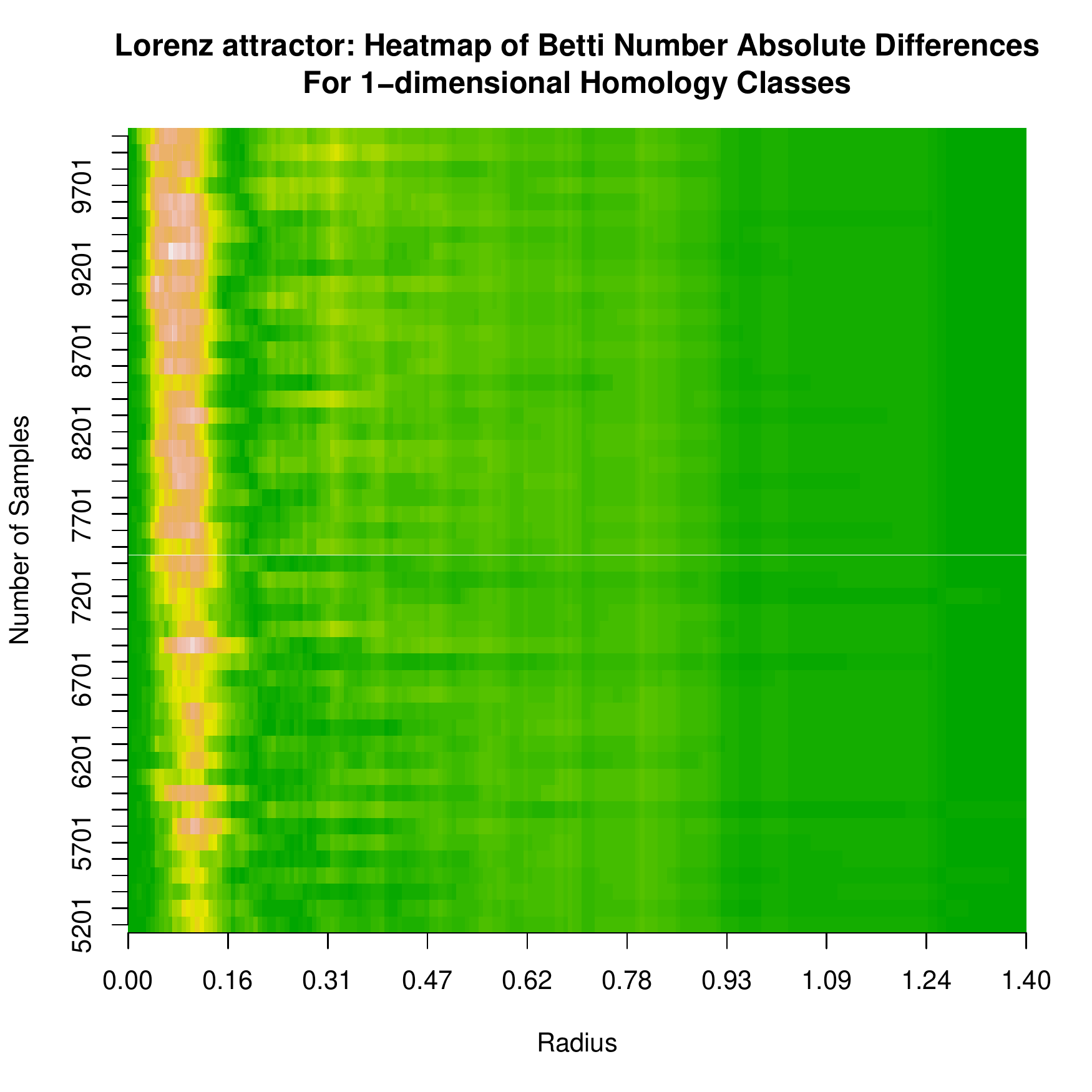}
\caption{The heatmap generated from the Lorenz attractor formed by values of the generalization measure $|\beta_1(\mathcal{X}_{r_i}) - \beta_1(\mathcal{X}_{r_i}^m)|$ along the insertion of samples and the increase of radius in the filtration.}
\label{fig:lorenz-heatmap-1}
\end{figure}

In addition, the visual inspection of the heatmap computed on the space $W_{r_i} \times r_i \times M$ (see Figure~\ref{fig:lorenz-heatmap-1}), leads to the conclusion that the value of $|\beta_1(\mathcal{X}_{r_i})-\beta_1(\mathcal{X}_{r_i}^m)|$ tends to increase with new data inclusions, whenever the radius is small, something also reported in prior experiments. Remember this is used to analyze the convergence of $W_{r_i}$ (Equation~\ref{eq:mult-criteria}), provided the inclusion of data, as the radius changes.

\subsection{Experiment III - R\"{o}ssler attractor}

The R\"{o}ssler attractor dataset was produced using the R\"{o}ssler system implemented in the package nonlinearTseries from the R Project for Statistical Computing~\footnote{Package nonlinearTseries is available at \url{https://cran.r-project.org/web/packages/nonlinearTseries}}. The parameters were set as follows: $a=0.2, b=0.2$, and $w=5.7$ with initial conditions given by $x_0=-2,y_0=-10$, and $z_0=0.2$. A total of $10,000$ observations were generated to compose the dataset illustrated in Figure~\ref{fig:rossler} (\ref{app:datasets}), varying time from $t_0=0$ to $t_f=50$ in steps of $0.005$.

\begin{figure}[htb!]
\centering
\includegraphics[scale=0.64]{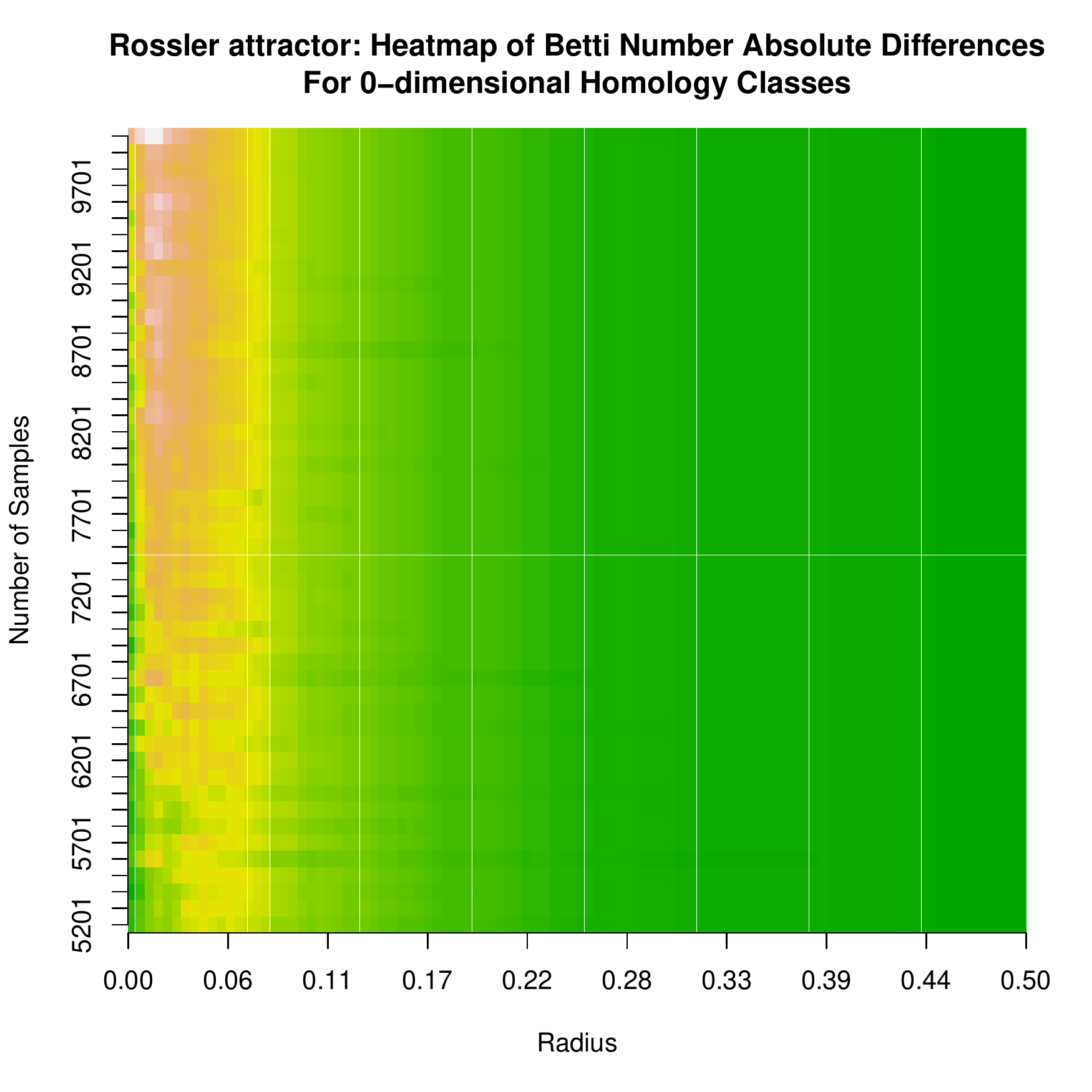}
\caption{The heatmap generated from the R\"{o}ssler attractor formed by values of the generalization measure $|\beta_0(\mathcal{X}_{r_i}) - \beta_0(\mathcal{X}_{r_i}^m)|$ along the insertion of samples and the increase of radius in the filtration.}
\label{fig:ross-heatmap-0}
\end{figure}

In the same approach as the one employed in previous experiments, we defined a filtration using the radius interval $[0,5)$ while employing Lazy Witness with $200$ landmarks in an attempt to produce simplicial complexes with degree at most one. Considering the $0$-dimensional complexes, the space $W_{r_i} \times r_i$ is illustrated in Figure~\ref{fig:ross-barcodes-0} (\ref{app:barcodes}), which was computed assuming $10,000$ values for $r_i$ in the range $[0, 5)$. This last figure confirms the behavior of previous experiments, such that in the interval $[1.5355,5)$ one has $|\beta_0(\mathcal{X}_{r_i})-\beta_0(\mathcal{X}_{r_i}^m)| = 0$. Such interval guarantees the identification of a single connected component and, as seen in Figure~\ref{fig:ross-heatmap-0}, for the first intervals, $|\beta_0(\mathcal{X}_{r_i})-\beta_0(\mathcal{X}_{r_i}^m)|$ increases along the number of samples, as expected.

\begin{figure}[htb!]
\centering
\includegraphics[scale=0.64]{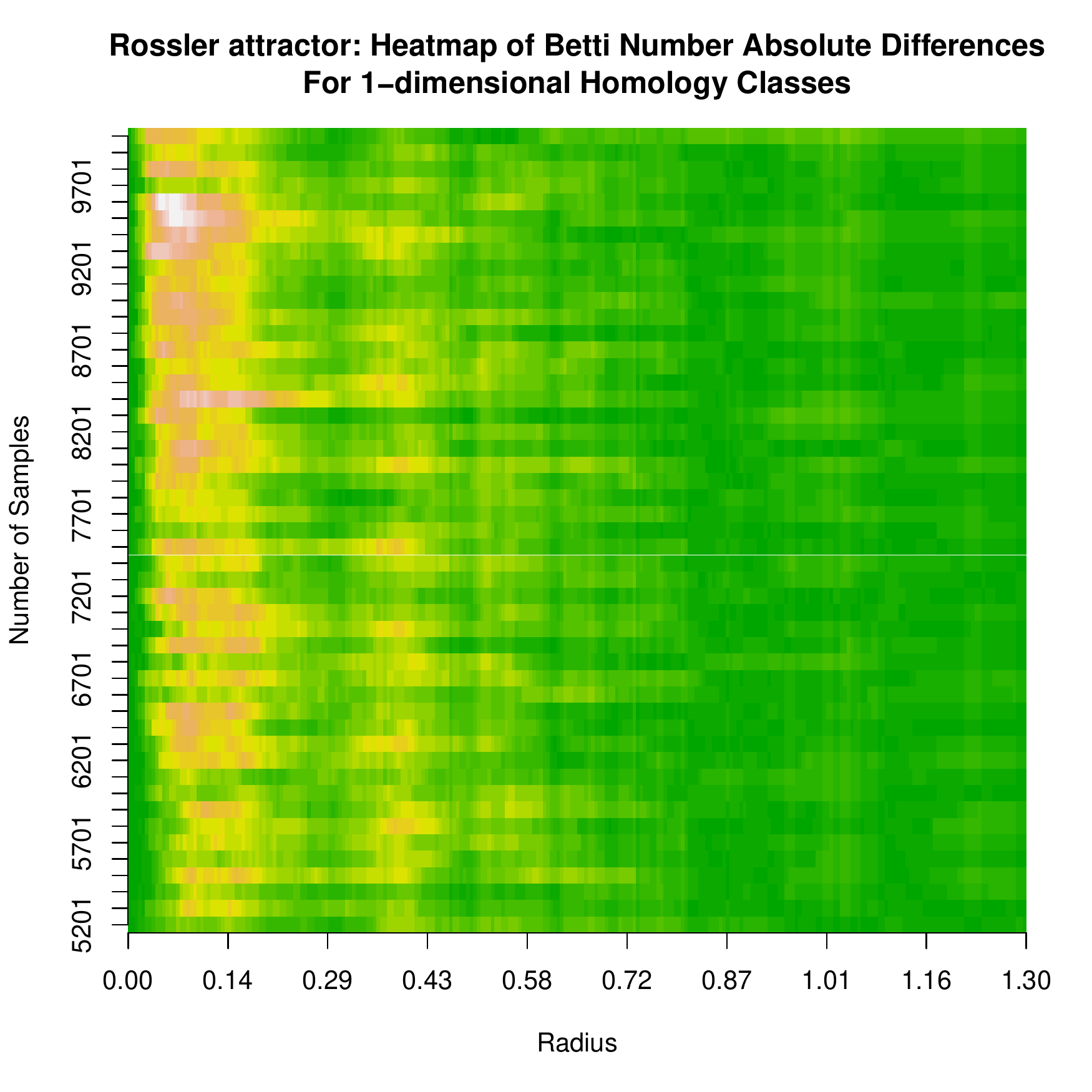}
\caption{The heatmap generated from the R\"{o}ssler attractor formed by values of the generalization measure $|\beta_1(\mathcal{X}_{r_i}) - \beta_1(\mathcal{X}_{r_i}^m)|$ along the insertion of samples and the increase of radius in the filtration.}
\label{fig:ross-heatmap-1}
\end{figure}

If we consider the study of one-dimensional homology classes, radii intervals $[0.5055,0.51),[0.5255,0.54),[0.5455,0.56)$, and $[4.4250,5)$ respect the generalization criteria proposed in Equation~\ref{eq:mult-criteria}. In this sense, the last interval represents the set of radii capable of identifying the attraction point of the R\"{o}ssler attractor, as illustrated in Figure~\ref{fig:ross-barcodes-1} (\ref{app:barcodes}). For instance, it is expected that any of the following intervals $[0.5055,0.51),[0.5255,0.54),[0.5455,0.56)$, and $[4.4250,5)$ will fail to produce consistent one-dimensional homology classes since $|\beta_1(\mathcal{X}_{r_i})-\beta_1(\mathcal{X}_{r_i}^m)| = 0$ is not guaranteed. Moreover, the heatmap (see Figure~\ref{fig:ross-heatmap-1}) for the space $W_{r_i} \times r_i \times M$ confirms the same behavior observed in previous experiments, showing an increase in $|\beta_1(\mathcal{X}_{r_i})-\beta_1(\mathcal{X}_{r_i}^m)|$ along the first radii intervals.

\subsection{Experiment IV - Mackey-Glass attractor}

The Mackey-Glass dataset was produced using the system implemented in the package frbs, also from the R Project for Statistical Computing~\footnote{Package nonlinearTseries is available at \url{https://cran.r-project.org/web/packages/nonlinearTseries}}, without the need of any parameter, as illustrated in Figure~\ref{fig:mackey} (\ref{app:datasets}). This fixed-size dataset only contains $1,000$ observations, therefore, in order to calculate the space $W_{r_i} \times r_i \times M$, the number of samples composing each dataset is given from $510$ to $1,000$ with steps of $20$ units. As before, the space $W_{r_i} \times r_i$ was defined by the difference $|\beta_p(\mathcal{X}_{r_i})-\beta_p(\mathcal{X}_{r_i}^m)|$ where $\beta(\mathcal{X}_{r_i})$ and $\beta(\mathcal{X}_{r_i}^m)$ are associated to, respectively, the datasets with $510$ and $1,000$ samples.

Subsequently, Lazy Witness was employed, with $200$ landmarks, to create filtrations of $0$-dimensional and one-dimensional simplicial complexes along the radius interval $[0,0.1)$. By studying the space $W_{r_i} \times r_i$, computed with $r_i$ assuming $10,000$ units from $0$ to $0.1$ (see Figure~\ref{fig:mackey-crit-0} in~\ref{app:barcodes}), the intervals $[0,0.0001)$ and $[0.0777,0.1)$ satisfy $|\beta_0(\mathcal{X}_{r_i})-\beta_0(\mathcal{X}_{r_i}^m)| = 0$, when $0$-dimensional homology classes are considered. The interval $[0,0.0001)$ guarantees convergence of the generalization measure to zero because it produces a simplicial complex which contains only the landmarks as connected components in the original and the perturbed dataset.

\begin{figure}[htb!]
\centering
\includegraphics[scale=0.64]{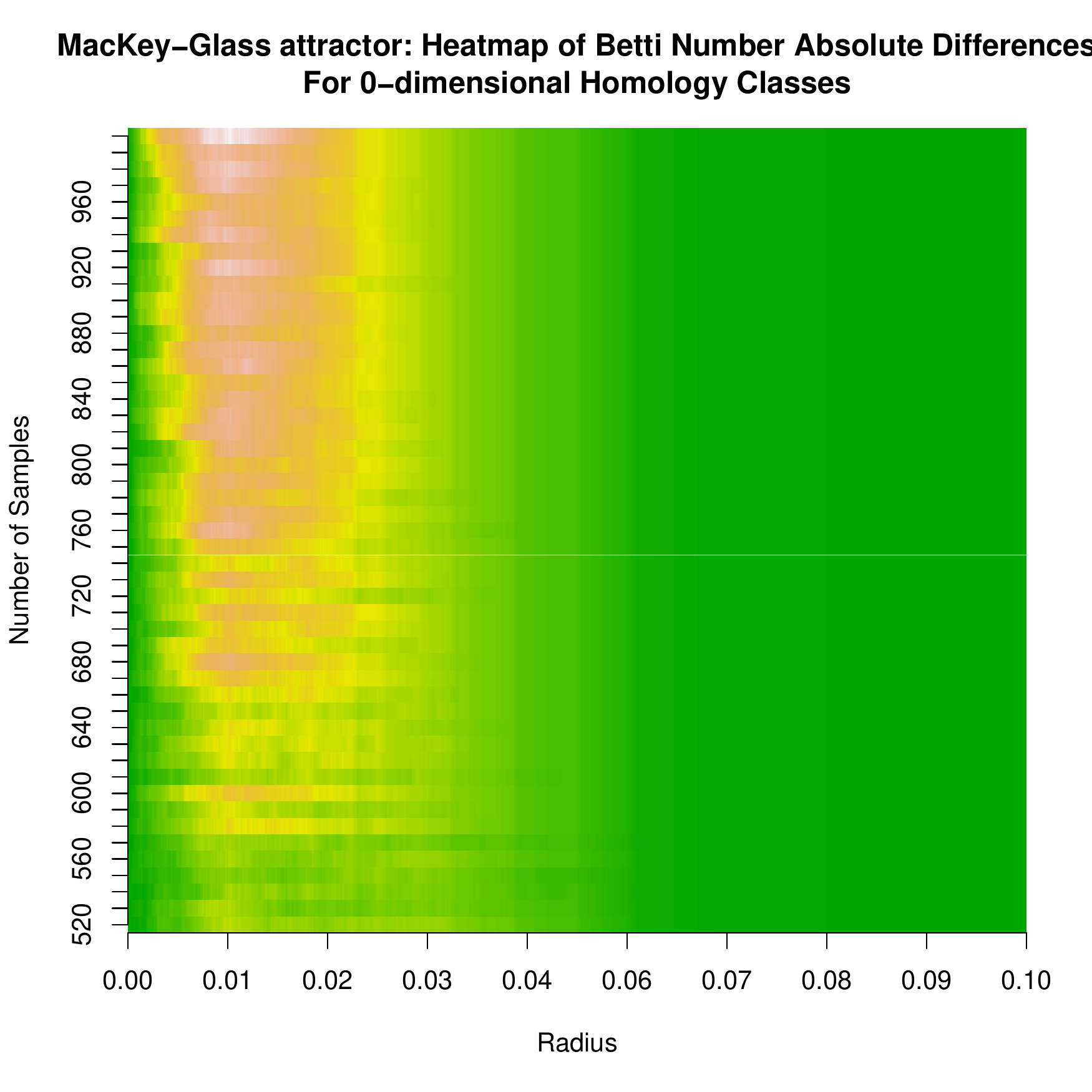}
\caption{The heatmap generated from the Mackey-Glass attractor formed by values of the generalization measure $|\beta_0(\mathcal{X}_{r_i}) - \beta_0(\mathcal{X}_{r_i}^m)|$ along the insertion of samples and the increase of radius in the filtration.}
\label{fig:heatmap-mackey-0}
\end{figure}

In addition, the interval $[0.0777,0.1)$ guarantees that only one connected component will be produced even when new data are included. As illustrated in Figure~\ref{fig:heatmap-mackey-0}, the first radii intervals produce an increase in the difference $|\beta_0(\mathcal{X}_{r_i})-\beta_0(\mathcal{X}_{r_i}^m)|$ as new observations are inserted in this dataset, as expected.

\begin{figure}[htb!]
\centering
\includegraphics[scale=0.64]{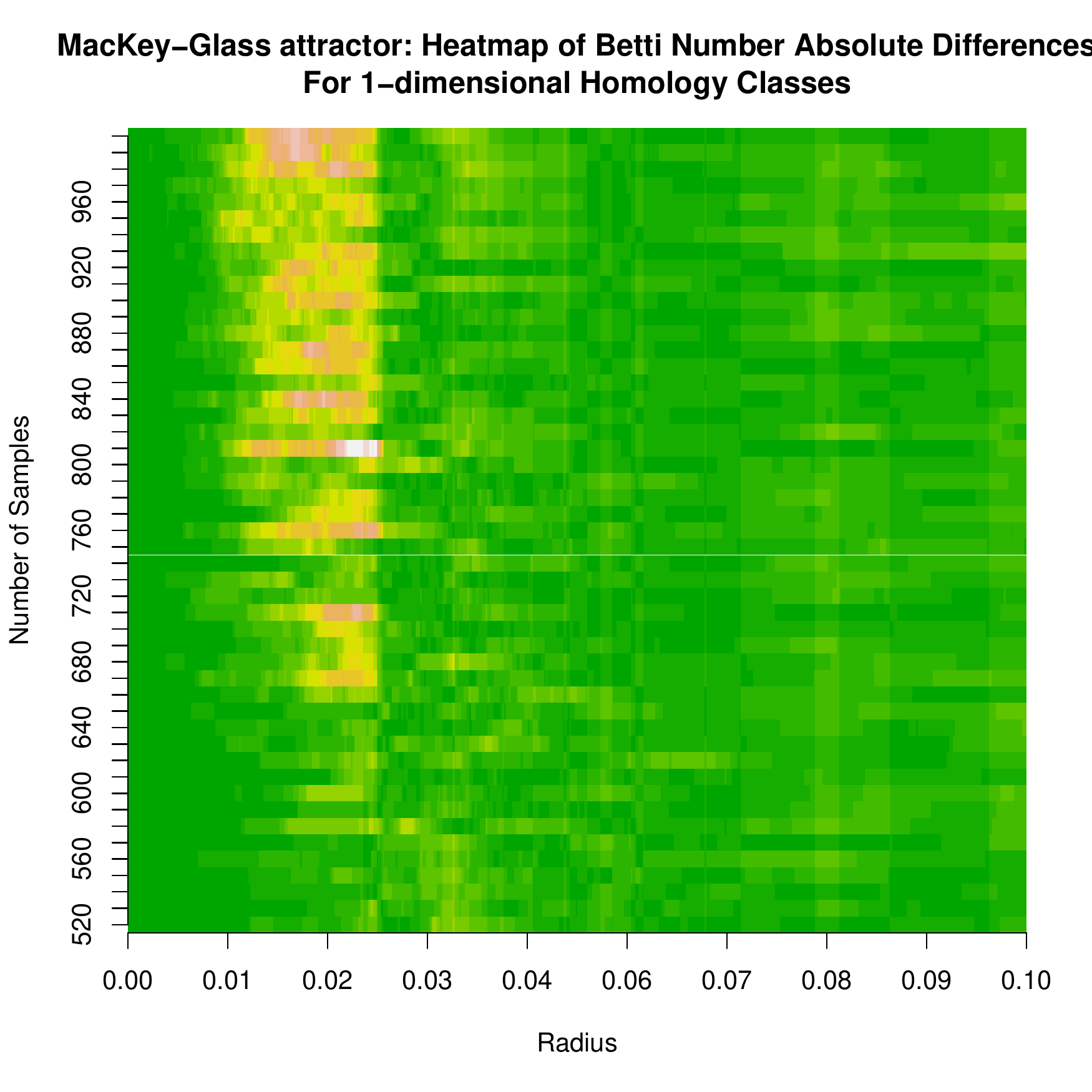}
\caption{The heatmap generated from the Mackey-Glass attractor formed by values of the generalization measure $|\beta_1(\mathcal{X}_{r_i}) - \beta_1(\mathcal{X}_{r_i}^m)|$ along the insertion of samples and the increase of radius in the filtration.}
\label{fig:heat-mackey-1}
\end{figure}

Considering one-dimensional homology classes, the space $W_{r_i} \times r_i$ (see Figure~\ref{fig:mackey-crit-1} in~\ref{app:barcodes}); it displays a very complex behavior which might be caused by insufficient data samples. Even though, as illustrated in Figure~\ref{fig:heat-mackey-1}, the first radii intervals still show an increase in $|\beta_1(\mathcal{X}_{r_i})-\beta_1(\mathcal{X}_{r_i}^m)|$ value along the inclusion of new data samples.

\section{Conclusion}\label{sec:conclusion}

Statistical Learning Theory (SLT) and Algorithm Stability (AS) provide theoretical frameworks to study the generalization of Supervised Machine Learning (SML) models based on risk/error functions~\citep{mello18}. However, those functions are not supported in the context of Unsupervised Machine Learning (UML) since labeled data is required, making difficult to analyze the convergence of Data Clustering (DC) models as well as to define a notion of generalization.

In this context, this paper introduces a proposal to assess learning generalization by measuring the similarity of DC models in terms of topological features given data perturbations. Such topological features are obtained from persistent homology~\citep{Edelsbrunner2002,edelsbrunner08} which supports the representation of hierarchical data organizations. Then, given a persistence mapping ${\bf f}_p^{i,j}:H_p(\mathcal{X}_i) \rightarrow H_p(\mathcal{X}_j)$, some dataset $X$ and a new dataset $X^m$ resultant from the inclusion of new $m$ samples in $X$, we prove that the absolute difference of $p$-th Betti-numbers $|\beta_p(\mathcal{X}_i)-\beta_p(\mathcal{X}_i^m)|$ indicates if the homology groups of the simplicial complexes of $\mathcal{X}_i$ and $\mathcal{X}_i^m$ are isomorphic, this is whether $p$-dimensional topological features are maintained even when the dataset $X$ is subject to inclusions. For instance, considering the DC problem on metric spaces, we verify whether the simplicial complex, associated to a partition and created using any radius $r_i > \epsilon$, $\mathcal{X}_i$ is isomorphic to new simplicial complexes built up after the inclusion of data following the same probability distribution.

We proved that, for $p=0$, the generalization measure $|\beta_p(\mathcal{X}_i)-\beta_p(\mathcal{X}_i^m)|$ presents a trade-off similar to the Bias-Variance dilemma considered in the context of the SLT~\citep{Vapnik95,luxburg09,mello18}. For instance, whenever the radius adopted in the metric space of some DC problem is small enough, $|\beta_p(\mathcal{X}_i)-\beta_p(\mathcal{X}_i^m)|$ will diverge as new data are included and, therefore, the homology groups of the simplicial complexes associated to $\mathcal{X}_i$ and $\mathcal{X}_i^m$ will not be isomorphic, leading to a phenomenon analogous to overfitting. Conversely, if the radius is big enough, $|\beta_p(\mathcal{X}_i)-\beta_p(\mathcal{X}_i^m)|$ will always equals zero and the homology groups of the simplicial complexes will always be isomorphic. The details of the topological structure will be lost though, causing a phenomenon similar to underfitting. This conclusion makes it evident that Kleinberg's richness axiom~\citep{kleinberg02} should be indeed relaxed when working with the DC problem, otherwise model consistency would not be ensured. We also prove that the rank of the $0$-homology group can be employed in order to provide lower and upper bounds for~\citet{carlsson10}'s consistency result and, finally, all experiments demonstrate the Coarse-Refinement dilemma, as discussed in Section~\ref{sec:exps}. 
\section*{Declaration of competing interest}{No competing interest.}
\section*{Acknowledgements}{This paper is based upon projects sponsored by FAPESP (S\~{a}o Paulo Research Foundation), CAPES (Coordination for the Improvement of Higher Education Personnel) and CNPq (National Counsel of Technological and Scientific Development), Brazil, under grants 2017/16548-6, PROEX-5881819/D, 302077/2017-0. Any opinions, findings, and conclusions or recommendations expressed in this material are those of the authors and/ do not necessarily reflect the views of FAPESP, CAPES, and CNPq.}

\bibliography{jmlr_coarse-refinement}
\bibliographystyle{elsarticle-num-names.bst}
\newpage
\appendix
\section{Notation}\label{app:notation}
\begin{multicols}{2}
\footnotesize\selectfont\noindent
$x,\dots,z$ (Lower case Latin letters from $x$ to $z$) : Variables;\\
$f,\dots,h$ (Lower case Latin letters from $f$ to $h$) : Functions;\\
$\alpha x$ or $\alpha d$ (Lower case Greek letters associated with a variable $x$ or function $d$) : Scalar multiplication;\\
$X$ (Upper case Latin letters) : Sets;\\
$\mathbb{R}^p,\mathbb{N}^p$ : Real and natural sets of dimension $p$;\\
$\Gamma$ : Partition of a set $S$ defined as in~\citet{kleinberg02}'s paper;\\
$(X,d)$ or $(X,d_X)$ : Metric space associated with a set $X$ of elements and a distance function $d$ (when such function is considered to be general) or $d_X$ (when such function is considered to be associated with $X$);\\
$(Z,\tau_Z)$ : The unknown underlying topological space (considered in the Data Clustering and Hierarchical Clustering problems);\\
$(X,\tau_X)$ : The topological space built on top of a dataset $X$;\\
$(\Omega,\tau_\Omega)$ : The unknown universal topological space (in which $(Z,\tau_Z)$ is included) associated with a Borel measure;\\
$\mathcal{N}(x)$ : Neighborhood of $x$;\\
$\eta$ : Map $\eta : x \mapsto \mathcal{N}(x)$;\\
$\mathcal{X}$ : Neighborhood topology;\\
$\sim$ : Equivalence relation;\\
$\sim_r$ : $r$-relation~\citep{carlsson10};\\
$\mathcal{N}_i(X)$ : A neighborhood topology associated with a level $i$ of a hierarchical clustering model;\\
$\eta_i$ : Map $\eta : x \mapsto \mathcal{N}_i(x)$;\\
$i,j$ : Indices related with persistence along topological coarse-refinement;\\
$t_i$ : Element of the pre-image of a tame function associated with an index $i$;\\
$k$ : An iterator;\\
$\Delta^p$ : Simplex of dimension $p$;\\
$\sigma$ : A singular $p$-simplex (a continuous map from a simplex to a topological space);\\
$\partial$ : The boundary operator;\\
$C_p$ : A $p$-dimensional simplicial complex;\\
$\Ima{\cdot}$ : Image of a map;\\
$\Ker{\cdot}$ : Kernel of a map;\\
$H_p$ : $p$-dimensional homology group; \\
$\mathbb{Z}^p$ : Abelian group of $p$-tuple of integers endowed with a direct sum;\\
$\mathbb{Q}$ : Field of rational numbers;\\
$\beta_p$ : $p$-th Betti-number;\\
$\mathbb{E}(\cdot)$ : Expected value;\\
$\mathcal{F}(X,\eta)$ : Filtration of neighborhood topological spaces associated with the map $\eta$ and dataset $X$;\\
$\mathbf{f}_p^{i,j}$ and $\mathbf{g}_p^{l,q}$ : $p$-dimensional persistence functions associated with, respectively, the coarse-refinement of a topological space and the inclusion of new data;\\
$\mathcal{X}_i$ and $\mathcal{X}_i^m$ : Neighborhood topological spaces associated, respectively, with $\mathcal{N}_i(X)$ and $\mathcal{N}_i(X^m)$;\\
$\iota_X$ and $\iota_\mathcal{N}$ : Inclusions $\mathcal{X}_i^{l < q} \subseteq \mathcal{X}_i^{q}$ and $\mathcal{X}^q_i \subseteq \mathcal{X}^q_j$ respectively;\\
$l,q$ : Indices related with the persistence along data inclusions;\\
$\mathcal{H}$ : Collection of abelian groups;\\
$\mu$ : A measure function;\\
$P_\mu$ : Probability function endowed with a measure function $\mu$;\\
$P_{\beta_p}$ : Probability function endowed with a $p$-th Betti measure;\\
$\mathcal{O}$ : Big-O notation;\\
$\mathbf{\Omega}$ : Big-$\Omega$ notation;\\
$M$,$N$ : Cardinalities;\\
$(X,u_d)$ : Ultrametric space;\\
$(X,d_X,\mu_X)$ : Measurable metric space (mm-space);\\
$\mathcal{S}_{mm}(X)$ : Measurable metric space (mm-space) associated with the dataset $X$;\\
$B(x,r)$ : Open ball of radius $r$ centered in $x$;\\
$\textnormal{supp}(\cdot)$ : Support of a measure function;\\
$d_H$ : Hausdorff distance;\\
$d_{GH}$ : Gromov-Hausdorff distance;\\
$\mathfrak{L}$ : ~\citet{carlsson10}'s adapted single-linkage function.
\end{multicols}

\newpage
\section{Dictionary of terms}\label{app:dict}
\begin{multicols}{2}
\footnotesize\selectfont\noindent
{\bf Stability}: Characteristic of the bounded absolute difference between a measure applied over a set of variables (typically random) and its perturbation (given by variable permutations, removals or inclusions);\\
{\bf Consistency}: Characteristic of an estimation which approximates its expected value as the number of random variables increases; \\
{\bf $p$-Homology consistency}: Consistency for the estimation of the rank of $p$-homology groups;\\
{\bf $p$-simplex}: $p$-dimensional generalization of triangles;\\
{\bf Singular $p$-simplex}: $p$-dimensional generalization of triangles built up from a continuous map applied over a topological space;\\
{\bf Singular $p$-chains}: Algebraic structure built up from singular $p$-simplex;\\
{\bf Boundary Operator}: Homomorphism applied over $p$-chains producing $p-1$-chains;\\
{\bf Chain complex}: Sequence of $p$-chains produced by the successive applications of the Boundary Operator;\\
{\bf Voids}: ``Empty space'' inside a topological space;\\
{\bf Holes}: ``Empty space'' which traverses a topological space;\\
{\bf Connected Component}: Maximal connected subsets which cannot be partitioned into two disjoint nonempty subsets;\\
{\bf Homology group}: Algebraic representation for Topological Space which considers its connected components, voids, and holes; \\
{\bf Homology class}: Algebraic representation of cycles in a Homology group;\\
{\bf Over-refinement}: Refinement of a topological space which produces unstable $p$-homology groups;\\\
{\bf Over-coarse}: Refinement of a topological space which vanishes relevant information such as number of connected components, holes, and voids;\\
{\bf Coarse-refinement Dilemma}: Dilemma in choosing an appropriate refinement in order to represent relevant topological features and avoid instability on the $p$-homology groups;\\
{\bf Dendrogram}: Representation of a hierarchical clustering model responsible for mapping a the set of grouped elements into the minimal radius which merged them all together;\\
{\bf Filtration}: Successive inclusions of topological spaces;\\
{\bf Persistence}: Interval in which a homology class persists throughout a filtration;\\
{\bf Persistence Function}: Function which maps a homology group of a included topological space into the homology group of a larger topological space;\\
{\bf $p$-th Betti-number}: The rank of a homology group;\\
{\bf Topological Data Analysis}: Scientific area responsible for studying topological properties of data points.\\
\end{multicols}

\newpage
\section{Proofs}\label{app:proofs}

\newproof{pot}{Proof of Lemma~\ref{lemma:DC-prob-conv}}
\begin{pot}
Let $X$ and $X^m$ be two dataset $X$ (such that $X^m = X \cup \{x'_1,\dots,x'_m\}$) with i.i.d. samples, $\mathcal{N}_i(\cdot)$ the neighborhood topology built on top of a points cloud and $\beta_p(\cdot)$ which defines a martingale along the data inclusion, the $p$-Betti-number calculated upon those neighborhood topologies. Following the Azuma's Inequality~\citep{azuma1967}
\begin{equation}
    P\left(|\beta_p(\mathcal{X}_i^m) - \beta_p(\mathcal{X}_i)| > \epsilon\right) \leq 2\exp\left(\frac{-\epsilon^2}{2\sum_{q=1}^{m} c_{q,p}^2}\right).
    \label{eq:mart-convergence}
\end{equation}

Note that, as $\beta_p(\cdot)$ assumes only integer numbers, all values of $c_{q,p}$ should be zero in order to ensure that the right factor in Inequality~\ref{eq:mart-convergence} becomes zero. Therefore, if Equation~\ref{eq:mult-criteria-2} which calculates the absolute difference between the $p$-th Betti-numbers associated to $\mathcal{X}_i$ and $\mathcal{X}_i^m$ holds, generalization is expected to occur
\begin{equation}
\forall m,\;|\beta_p(\mathcal{X}_i^m) - \beta_p(\mathcal{X}_i)| = 0.
\label{eq:mult-criteria-2}
\end{equation}

Considering Equation~\ref{eq:mart-mexp}, the sequence $\beta_p(\mathcal{X}_i^q) - \beta_p(\mathcal{X}_i)$ with $q=1,\dots,m$ also forms a martingale
\begin{equation*}
\begin{array}{rcl}
        \mathbb{E}\left[\beta_p(\mathcal{X}_i^q) - \beta_p(\mathcal{X}_i)\right]  & = &  \mathbb{E}\left[\beta_p(\mathcal{X}_i^q)\right] - \beta_p(\mathcal{X}_i)\\
        & = & \beta_p(\mathcal{N}_i(X^{q-1})) - \beta_p(\mathcal{X}_i).
    \end{array}
\end{equation*}
Therefore, in terms of the average $p$-th Betti-number differences, if $|\beta_p(\mathcal{X}_i^q) - \beta_p(\mathcal{X}_i)| < c_{q,p}$, probabilistic convergence is defined as
\begin{equation}
    P\left(\sum_{q=1}^{m} |\beta_p(\mathcal{X}_i^q) - \beta_p(\mathcal{X}_i)| > m\epsilon\right) \leq 2\exp\left(\frac{-m^2\epsilon^2}{2\sum_{q=1}^{m} c_{q,p}^2}\right),
\end{equation}
\end{pot}

\newproof{pot2}{Proof of Corollary~\ref{cor:DC-max_c}}
\begin{pot2}
Following from Inequality~\ref{eq:crit-prob-convergence} of Lemma~\ref{lemma:DC-prob-conv} can be written as
\begin{equation*}
    P\left(\sum_{q=1}^{m} |\beta_p(\mathcal{X}_i^q) - \beta_p(\mathcal{X}_i)| \leq m\epsilon\right) > 2\exp\left(\frac{-m^2\epsilon^2}{2\sum_{q=1}^{m} c_{q,p}^2}\right),
\end{equation*}
which, substituting $ P\left(\sum_{q=1}^{m} |\beta_p(\mathcal{X}_i^q) - \beta_p(\mathcal{X}_i)| \leq m\epsilon\right) = \delta$, results in
\begin{equation}
    \delta > 2\exp\left(\frac{-m^2\epsilon^2}{2\sum_{q=1}^{m} c_{q,p}^2}\right).
    \label{eq:mart-delta}
\end{equation}
Let $\overline{c}_p$ be the maximum value of $(c_{q,p})_{q=1\dots,m}$; then, it follows from Inequality~\ref{eq:mart-delta} that
\begin{equation}
    \delta > 2\exp\left(\frac{-m^2\epsilon^2}{2\sum_{q=1}^{m} c_{q,p}^2}\right) \geq 2\exp\left(\frac{-m^2\epsilon^2}{2m \overline{c}^2_p}\right).
    \label{eq:mart-delta-max}
\end{equation}
So,
\begin{equation*}
    \begin{array}{rcl}
        \delta                  & > & 2\exp\left(\frac{-m^2\epsilon^2}{2m\overline{c}^2_p}\right) \\[7pt]
        \ln{\delta}             & > & \ln{2} - \frac{m\epsilon^2}{2\overline{c}^2_p} \\[7pt]
        \ln{2} - \ln{\delta}    & < & \frac{m\epsilon^2}{2\overline{c}^2_p} \\[7pt]
        \overline{c}_p          & < & \sqrt{\frac{m\epsilon^2}{2(\ln{2}-\ln{\delta}})}.
    \end{array}
\end{equation*}
Since $\delta=1$ implies in consistency, so that
\begin{equation}
    \overline{c}_p < \epsilon \sqrt{m/2\ln{2}}
\end{equation}
guarantees it.
\end{pot2}

\newproof{pot3}{Proof of Theorem~\ref{theo:HC-conv}}
\begin{pot3}
Defining $\overline{\Delta\beta}_{p,i}=\sum_{q=1}^{m} |\beta_p(\mathcal{X}_i^q) - \beta_p(\mathcal{X}_i)|$:
\begin{equation*}
\begin{array}{rcl}
    P\left(\sup_{i \in \mathbb R} \overline{\Delta\beta}_{p,i} > m\epsilon\right) & = & P\left(\bigvee_{i \in \mathbb R} (\overline{\Delta\beta}_{p,i}) > m\epsilon\right) \\[7pt]
    & \leq & \sum_{i \in \mathbb R} P\left(\overline{\Delta\beta}_{p,i} > m\epsilon\right) \\[7pt]
    & \leq & 2M \exp\left(\frac{-m\epsilon^2}{2\tilde{c}_p^2}\right),
\end{array}
\end{equation*}
where $M$ is the amount of critical points a tame function $f$ and $\tilde{c}_p = \max_{i \in \mathbb R} \overline{c}_{p,i}$.
\end{pot3}

\newpage
\section{Dataset images}\label{app:datasets}

This appendix contains the illustrations of all datasets employed on the experiments of this paper, which are: the bidimensional torus, the Crescent Moon dataset, the Lorenz Attractor, the R\"{o}ssler Attractor and the MacKey-Glass Attractor.

\begin{figure}[htb!]
\centering
\includegraphics[scale=0.64]{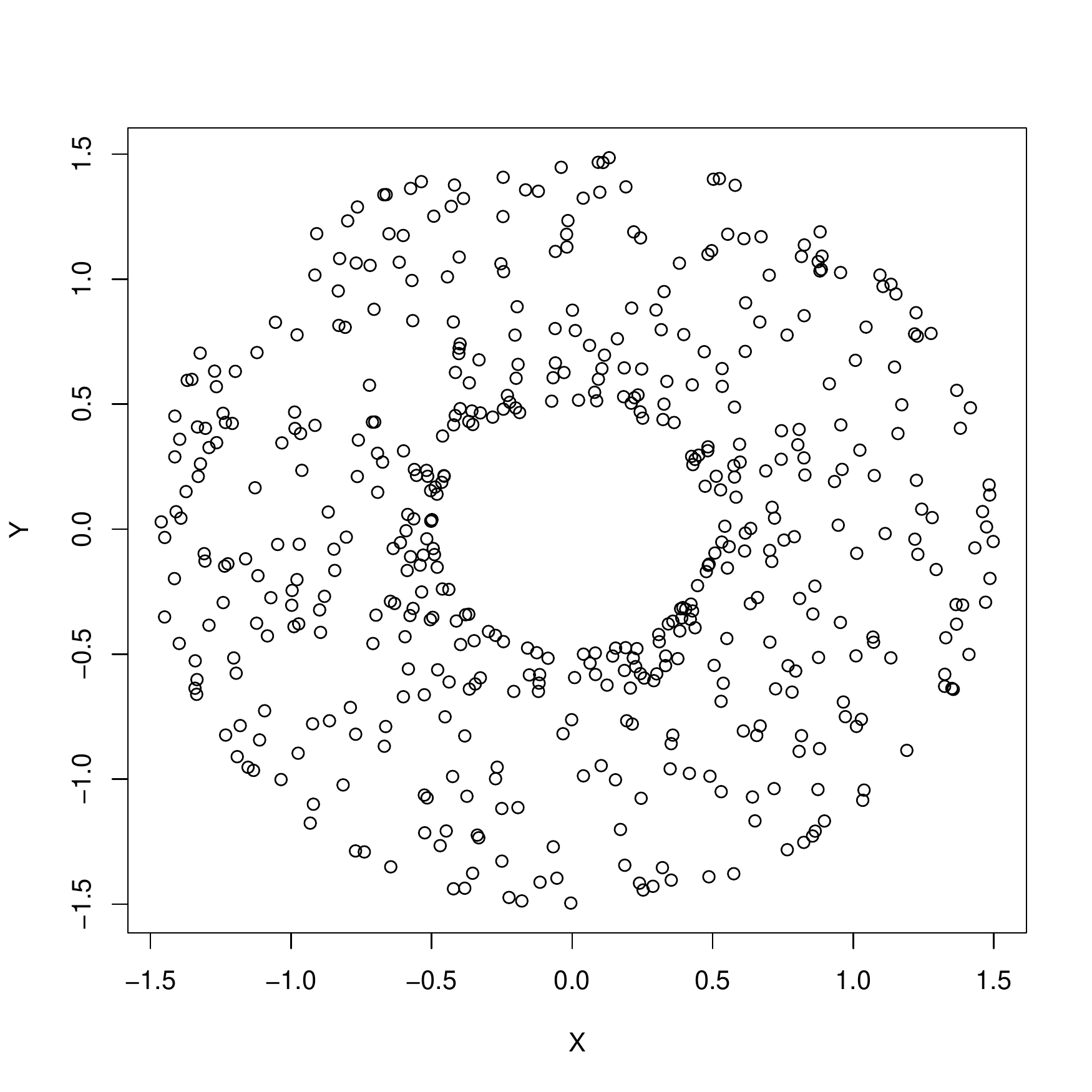}
\caption{A two-dimensional torus formed by a point cloud of $500$ samples generated using Equation~\ref{eq:tor} with parameters set to $R=1,r=0.5,\phi \in [0,2\pi]$, and $\theta \in [0,2\pi]$.}
\label{fig:data-torus}
\end{figure}
\newpage
\begin{figure}[htb!]
\centering
\includegraphics[scale=0.64]{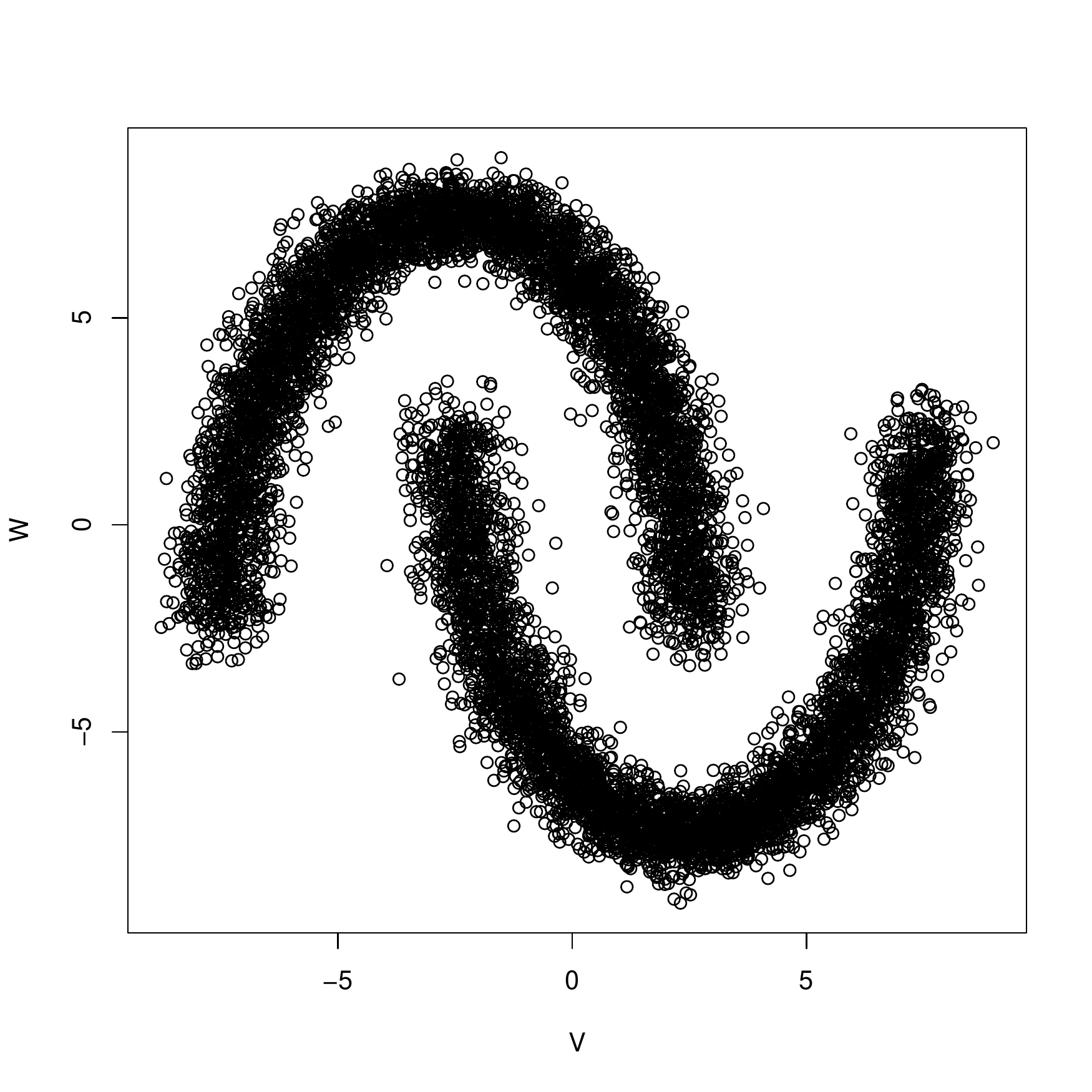}
\caption{The Crescent Moon dataset produced by the function {\it generateCrescentMoon} of the RSSL package from the R Project for Statistical Computing, adopting the following parameters: $n=5,000,d=2$ and $\sigma=0.5$.}
\label{fig:crescmoon}
\end{figure}
\newpage
\begin{figure}[htb!]
\centering
\includegraphics[scale=0.64]{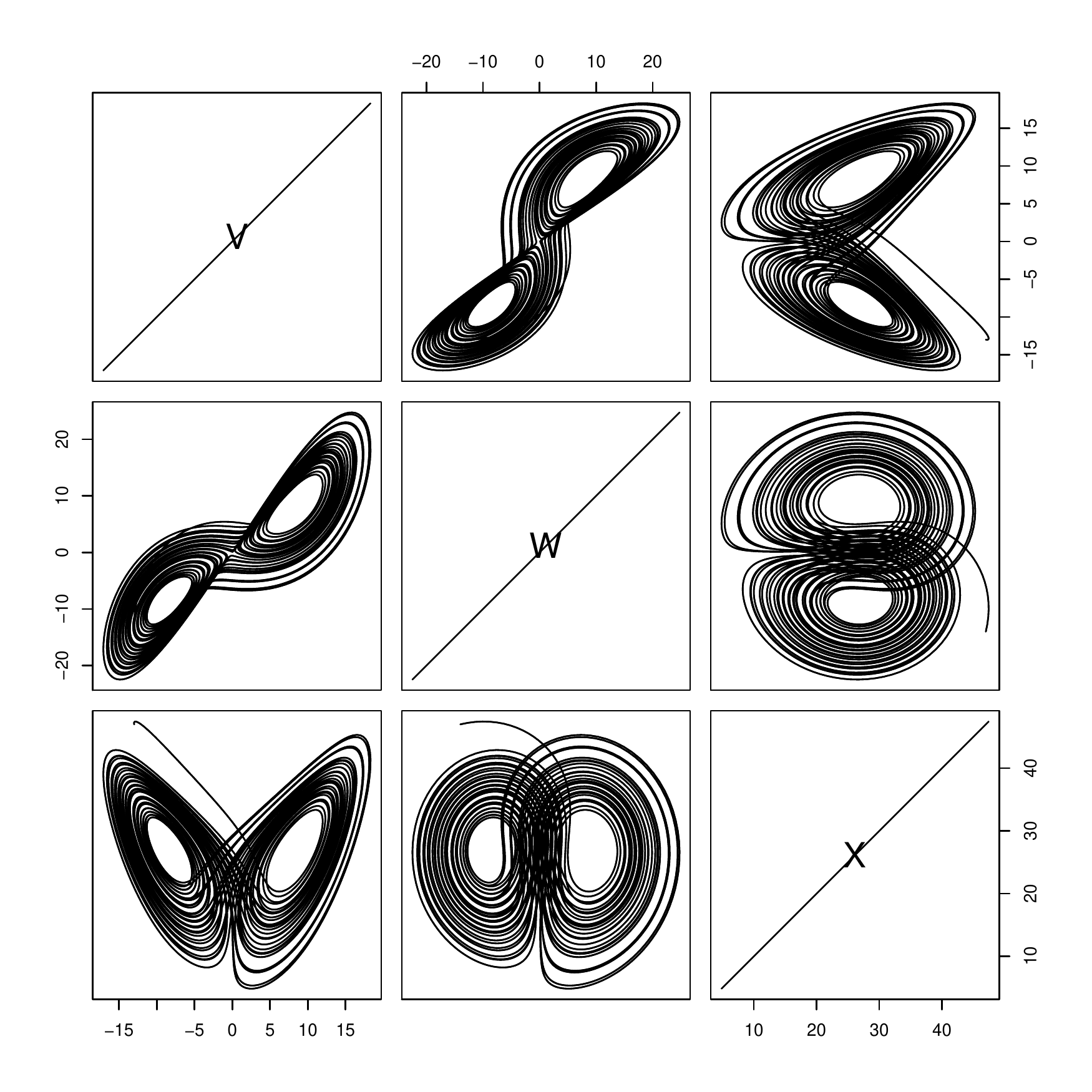}
\caption{The Lorenz attractor generated by the {\it lorenz} function of the package nonlinearTseries (from the R Project for Statistical Computing) with the following parameters: $\sigma=10,\beta=8/3$ and $\rho=28$ with initial conditions given by $x_0=-13,y_0=-14$ and $z=47$.}
\label{fig:lorenz}
\end{figure}
\newpage
\begin{figure}[htb!]
\centering
\includegraphics[scale=0.64]{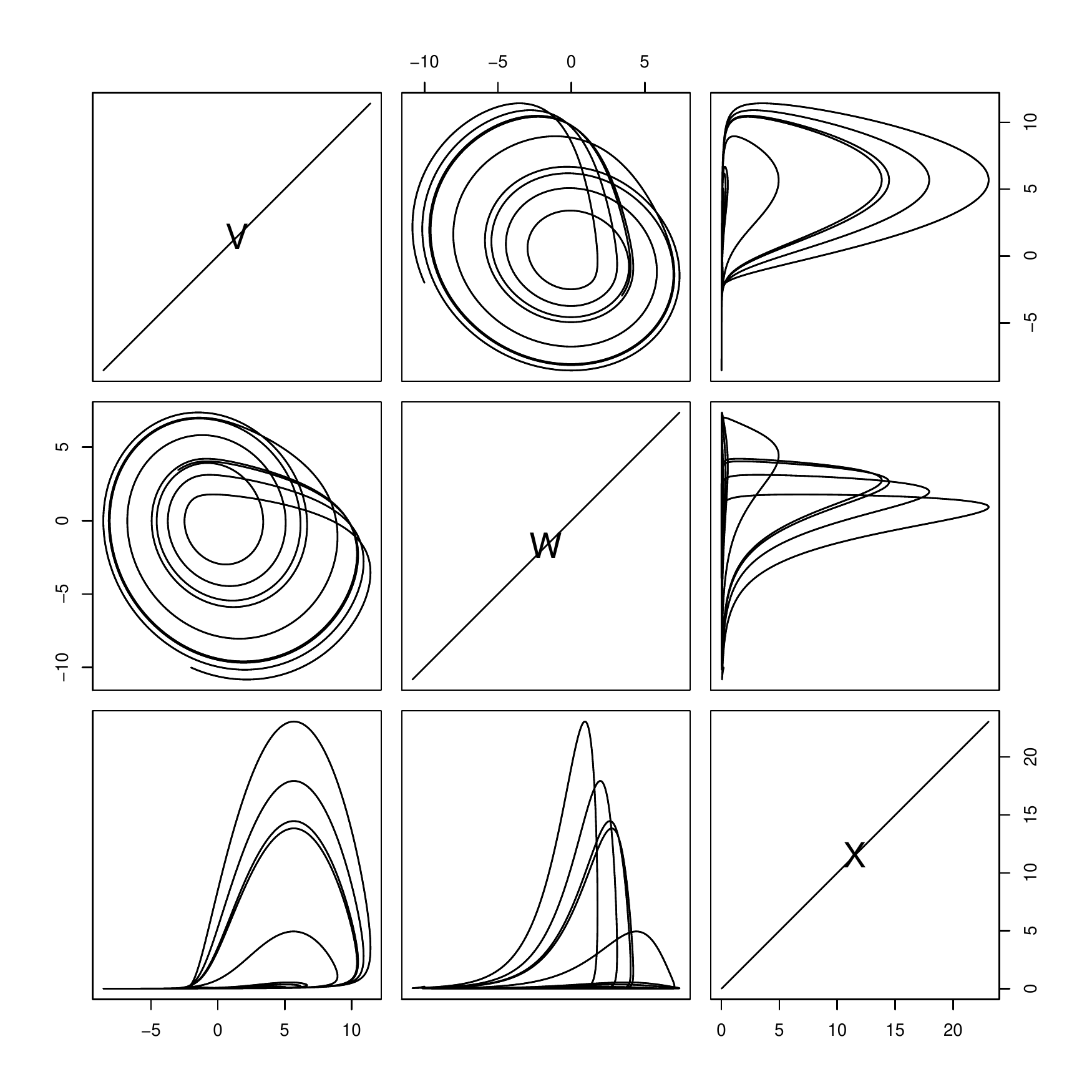}
\caption{The R\"{o}ssler attractor generated by the function {\it rossler} (implemented in package nonlinearTseries from the R Project for Statistical Computing) with the following parameters: $a=0.2, b=0.2$ and $w=5.7$ with initial conditions given by $x_0=-2,y_0=-10$ and $z_0=0.2$.}
\label{fig:rossler}
\end{figure}
\newpage
\begin{figure}[htb!]
\centering
\includegraphics[scale=0.64]{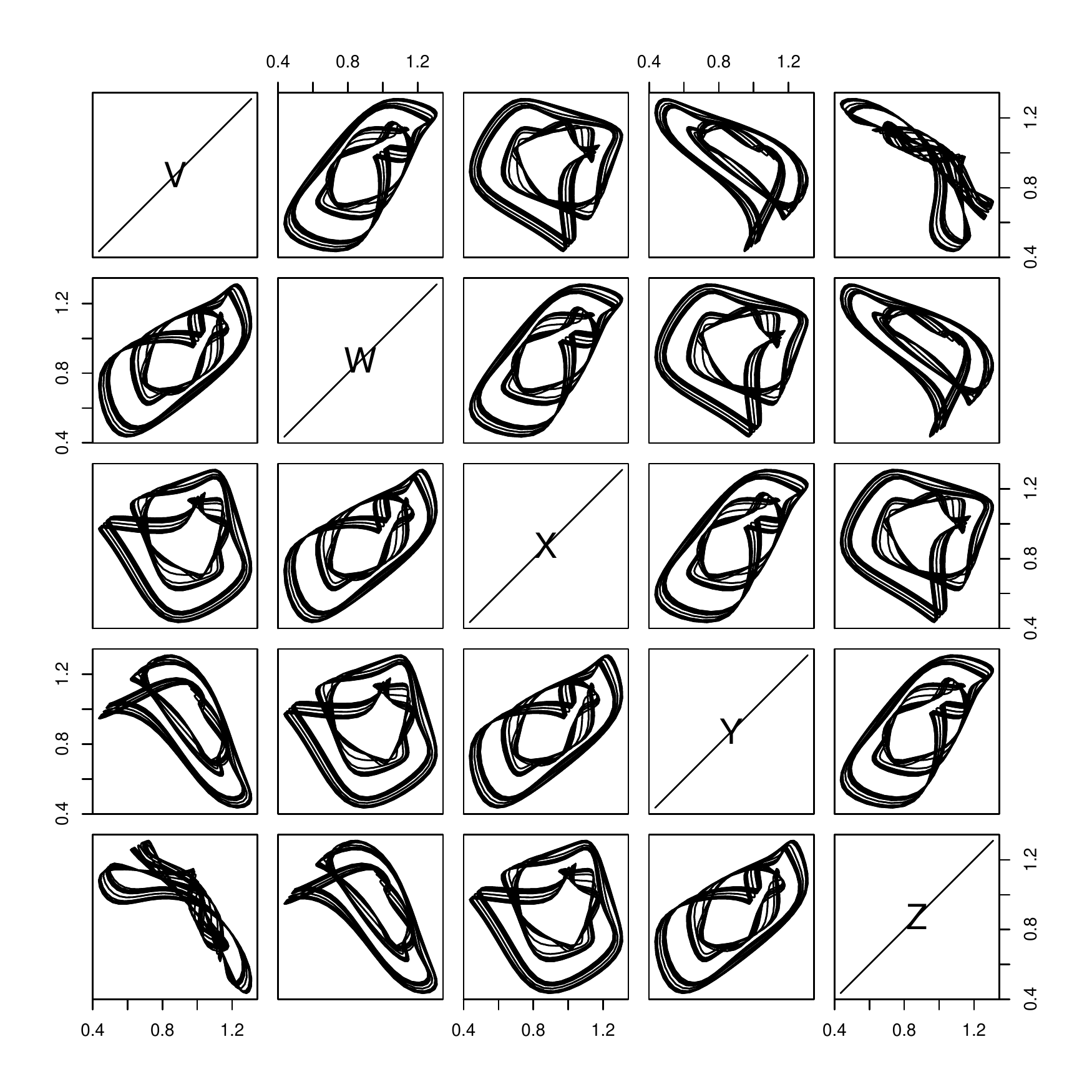}
\caption{The Mackey-Glass attractor available with the package frbs (from the R Project for Statistical Computing).}
\label{fig:mackey}
\end{figure}

\newpage
\section{Barcode Plots}~\label{app:barcodes}

This appendix contains the barcode plots produced by the datasets: Crescent Moon, Lorenz Attractor, R\"{o}ssler Attractor and MacKey-Glass Attractor. They are produced considering the dataset with and without perturbation and the generalization measure $|\beta_p(\mathcal{X}_{r_i}) - \beta_p(\mathcal{X}_{r_i}^m)|$ is also considered in the illustrations.

\begin{figure}[htb!]
\centering
\includegraphics[scale=0.64]{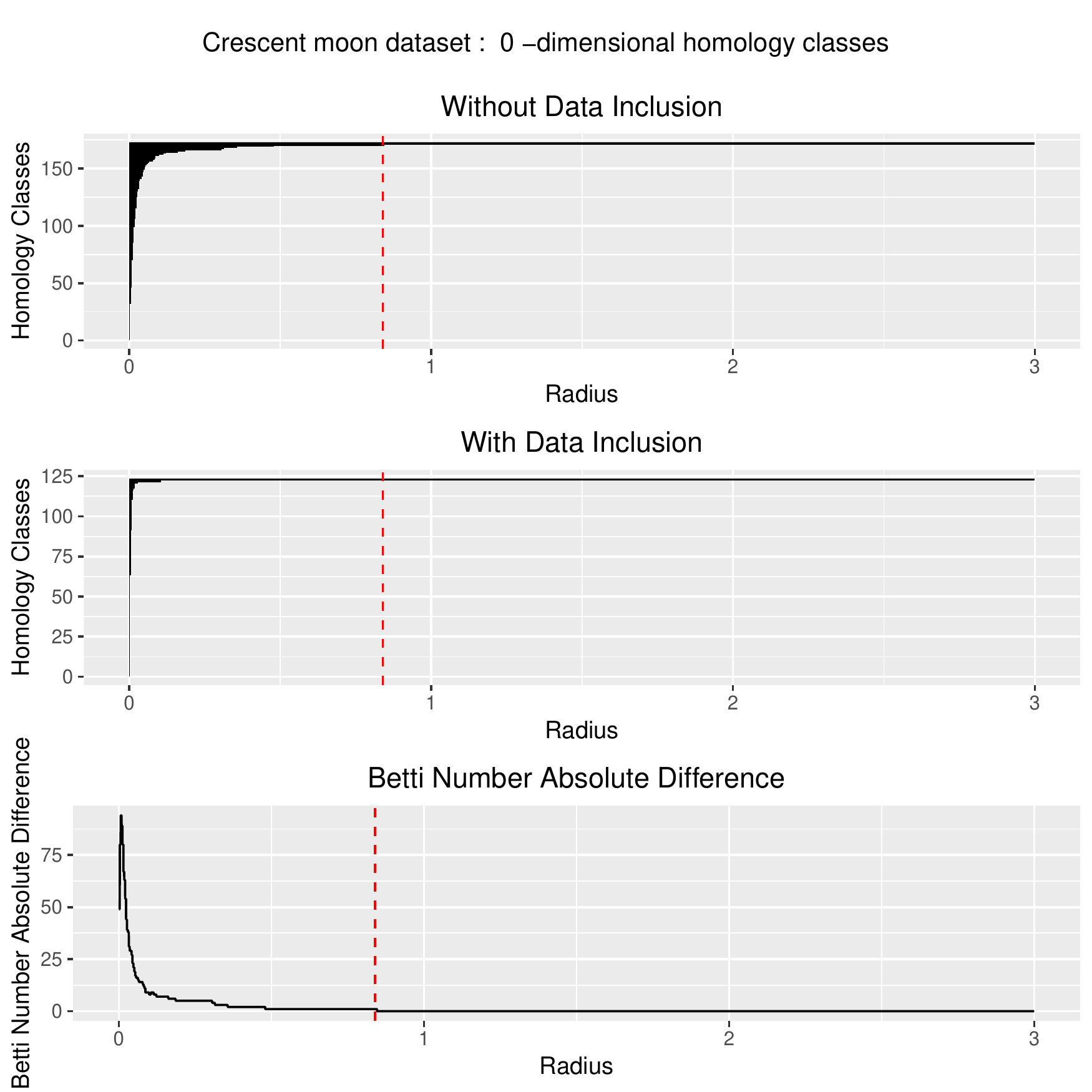}
\caption{Graphs produced from $0$-dimensional homology classes which correspond to, from top to bottom: i) Barcode plots generated using the Crescent Moon dataset with $5,100$ samples; ii) Barcode plots generated using the perturbed dataset with $10,000$ samples; and, finally, iii) the values for the generalization measure $|\beta_0(\mathcal{X}_{r_i}) - \beta_0(\mathcal{X}_{r_i}^m)|$. The red-dashed lines mark the initial value of the intervals which ensure $|\beta_0(\mathcal{X}_{r_i}) - \beta_0(\mathcal{X}_{r_i}^m)| = 0$.}
\label{fig:moon-barcodes-0}
\end{figure}

\newpage
\begin{figure}[htb!]
\centering
\includegraphics[scale=0.64]{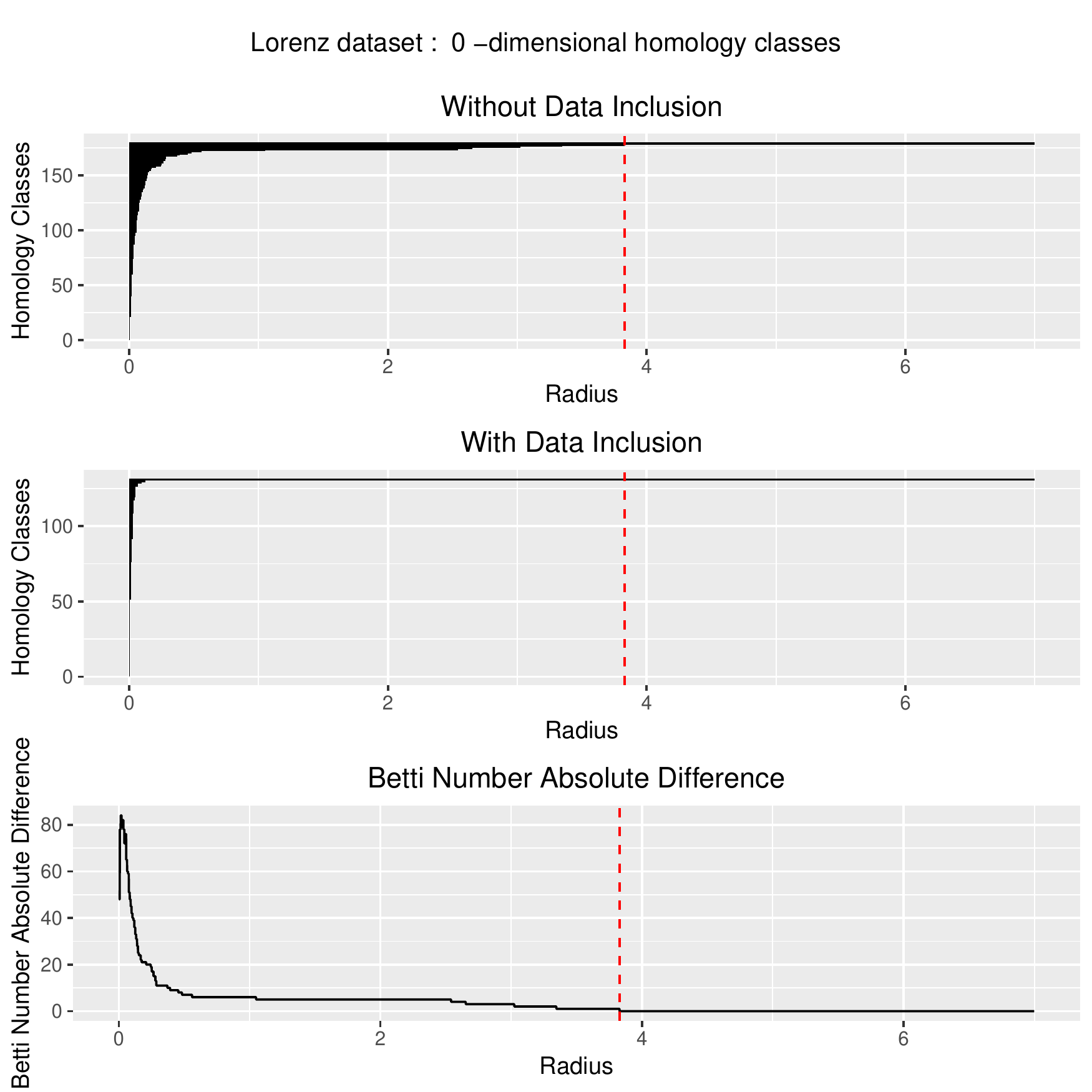}
\caption{Graphs produced from $0$-dimensional homology classes which correspond to, from top to bottom: i) Barcode plots generated using the Lorenz attractor experiment with $5,101$ samples; ii) Barcode plots generated using the perturbed dataset with $10,001$ samples; and, finally, iii) the values for the generalization measure $|\beta_0(\mathcal{X}_{r_i}) - \beta_0(\mathcal{X}_{r_i}^m)|$. The red-dashed lines mark the initial value of the intervals which ensure $|\beta_0(\mathcal{X}_{r_i}) - \beta_0(\mathcal{X}_{r_i}^m)| = 0$.}
\label{fig:lorenz-barcodes-0}
\end{figure}

\newpage
\begin{figure}[htb!]
\centering
\includegraphics[scale=0.64]{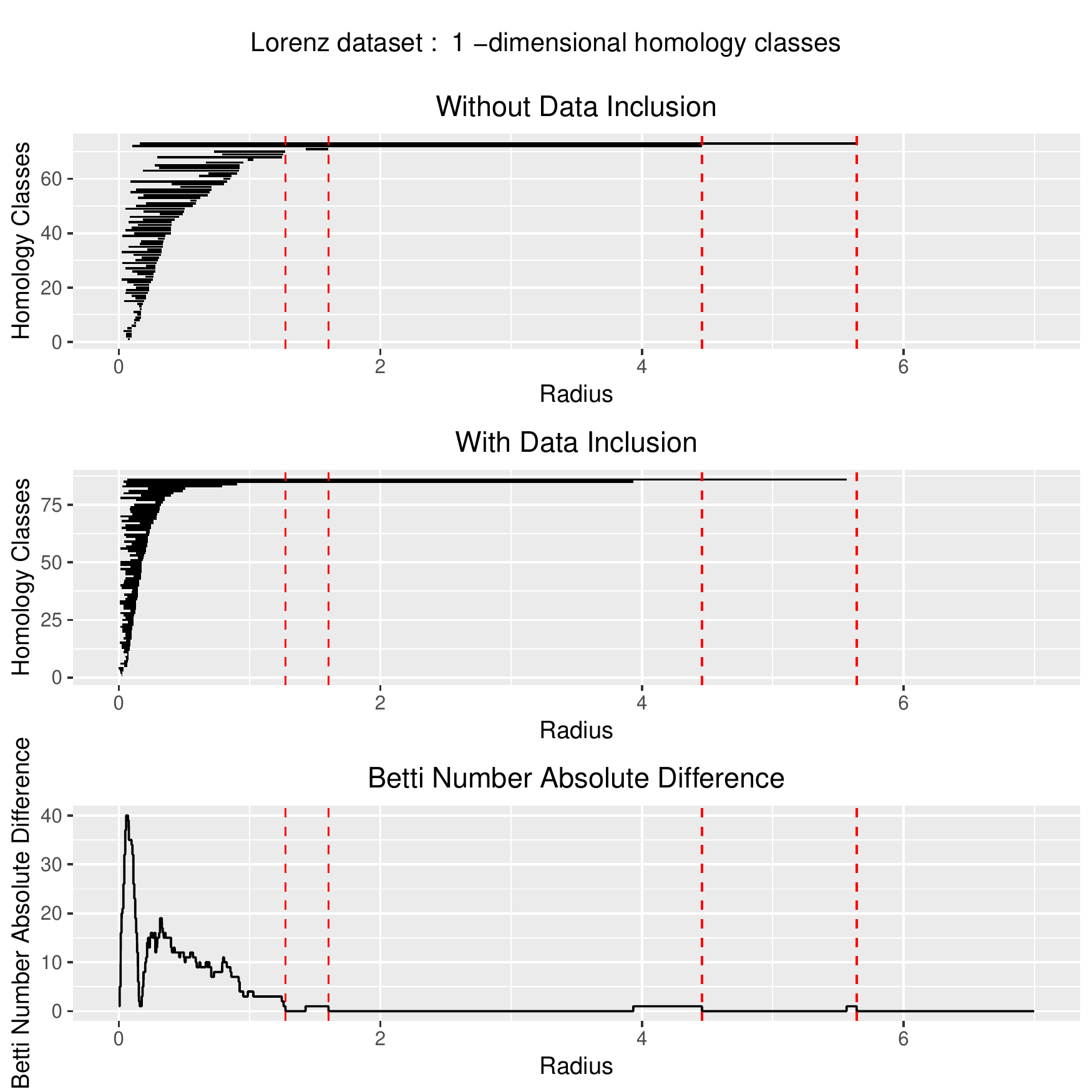}
\caption{Graphs produced from one-dimensional homology classes which correspond to, from top to bottom: i) Barcode plots generated using the Lorenz attractor experiment with $5,101$ samples; ii) Barcode plots generated using the perturbed dataset with $10,001$ samples; and, finally, iii) the values for the generalization measure $|\beta_1(\mathcal{X}_{r_i}) - \beta_1(\mathcal{X}_{r_i}^m)|$. The red-dashed lines mark the initial value of the intervals which ensure $|\beta_1(\mathcal{X}_{r_i}) - \beta_1(\mathcal{X}_{r_i}^m)| = 0$.}
\label{fig:lorenz-barcodes-1}
\end{figure}

\newpage
\begin{figure}[htb!]
\centering
\includegraphics[scale=0.64]{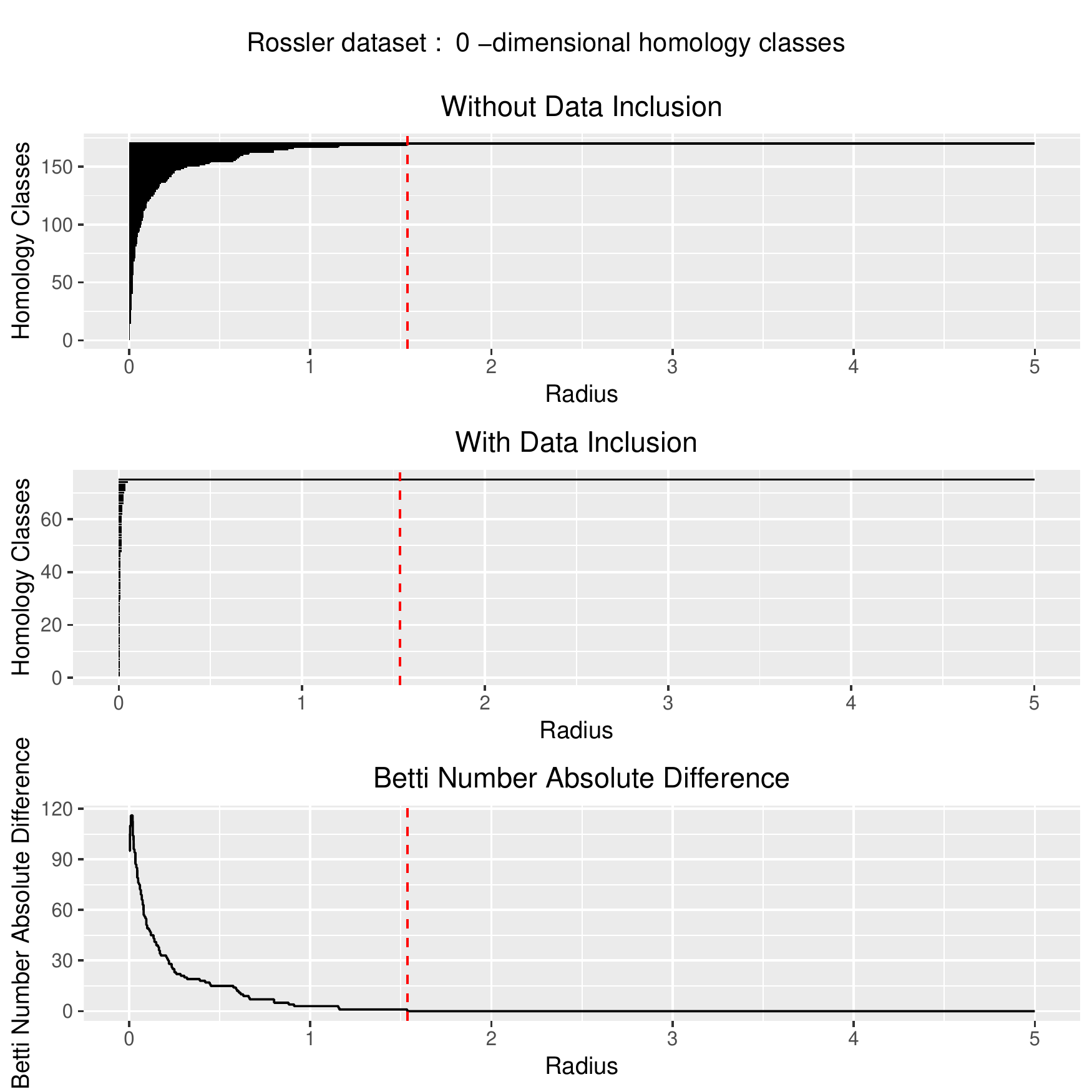}
\caption{Graphs produced from $0$-dimensional homology classes which correspond to, from top to bottom: i) Barcode plots generated using the R\"{o}ssler attractor experiment with $5,101$ samples; ii) Barcode plots generated using the perturbed dataset with $10,001$ samples; and, finally, iii) the values for the generalization measure $|\beta_0(\mathcal{X}_{r_i}) - \beta_0(\mathcal{X}_{r_i}^m)|$. The red-dashed lines mark the initial value of the intervals which ensure $|\beta_0(\mathcal{X}_{r_i}) - \beta_0(\mathcal{X}_{r_i}^m)| = 0$.}
\label{fig:ross-barcodes-0}
\end{figure}

\newpage
\begin{figure}[htb!]
\centering
\includegraphics[scale=0.64]{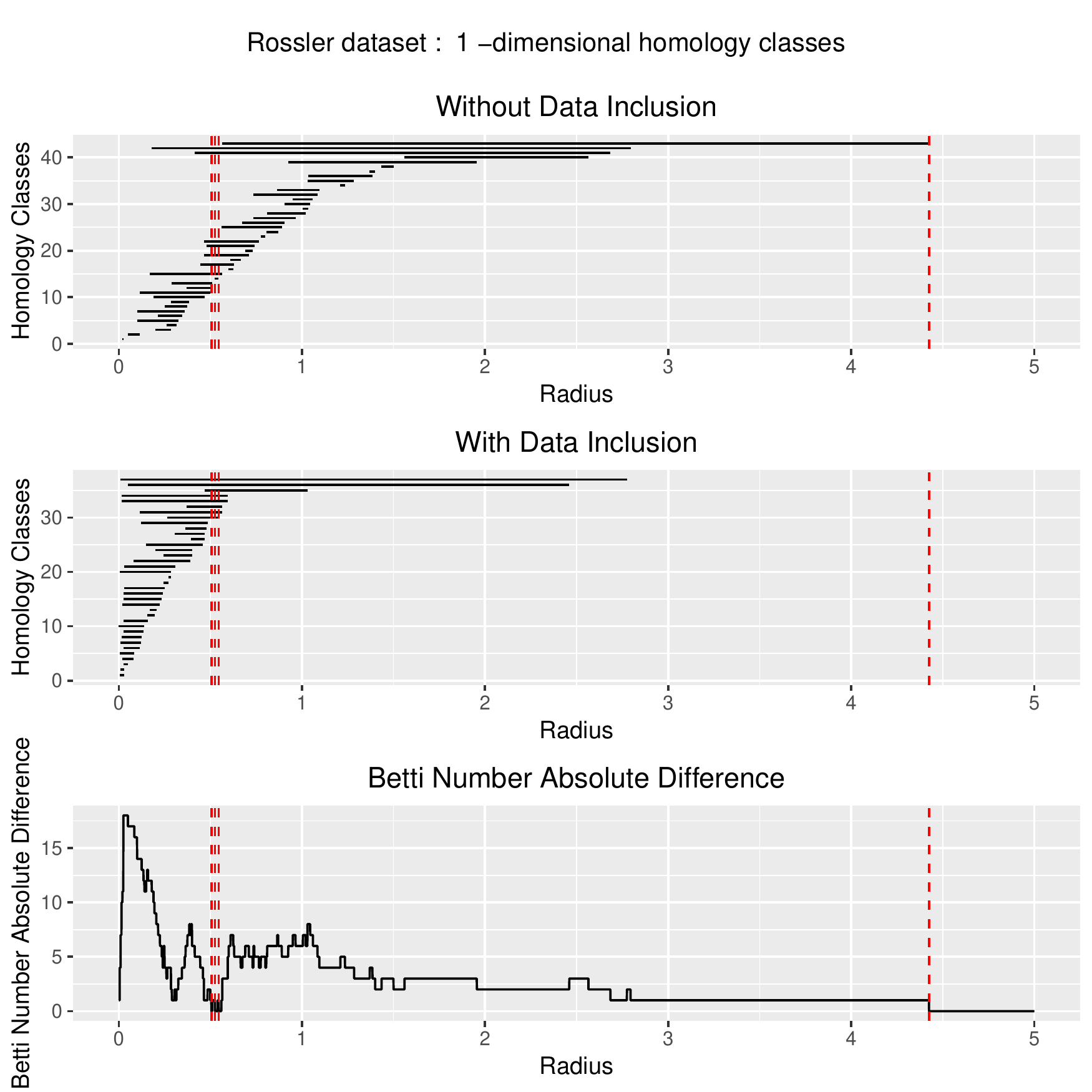}
\caption{Graphs produced from one-dimensional homology classes which correspond to, from top to bottom: i) Barcode plots generated using the R\"{o}ssler attractor prior described with $5,101$ samples; ii) Barcode plots generated using the perturbed dataset with $10,001$ samples; and, finally, iii) the values for the generalization measure $|\beta_1(\mathcal{X}_{r_i}) - \beta_1(\mathcal{X}_{r_i}^m)|$. The red-dashed lines mark the initial value of the intervals which ensure $|\beta_1(\mathcal{X}_{r_i}) - \beta_1(\mathcal{X}_{r_i}^m)| = 0$.}
\label{fig:ross-barcodes-1}
\end{figure}

\newpage
\begin{figure}[htb!]
\centering
\includegraphics[scale=0.64]{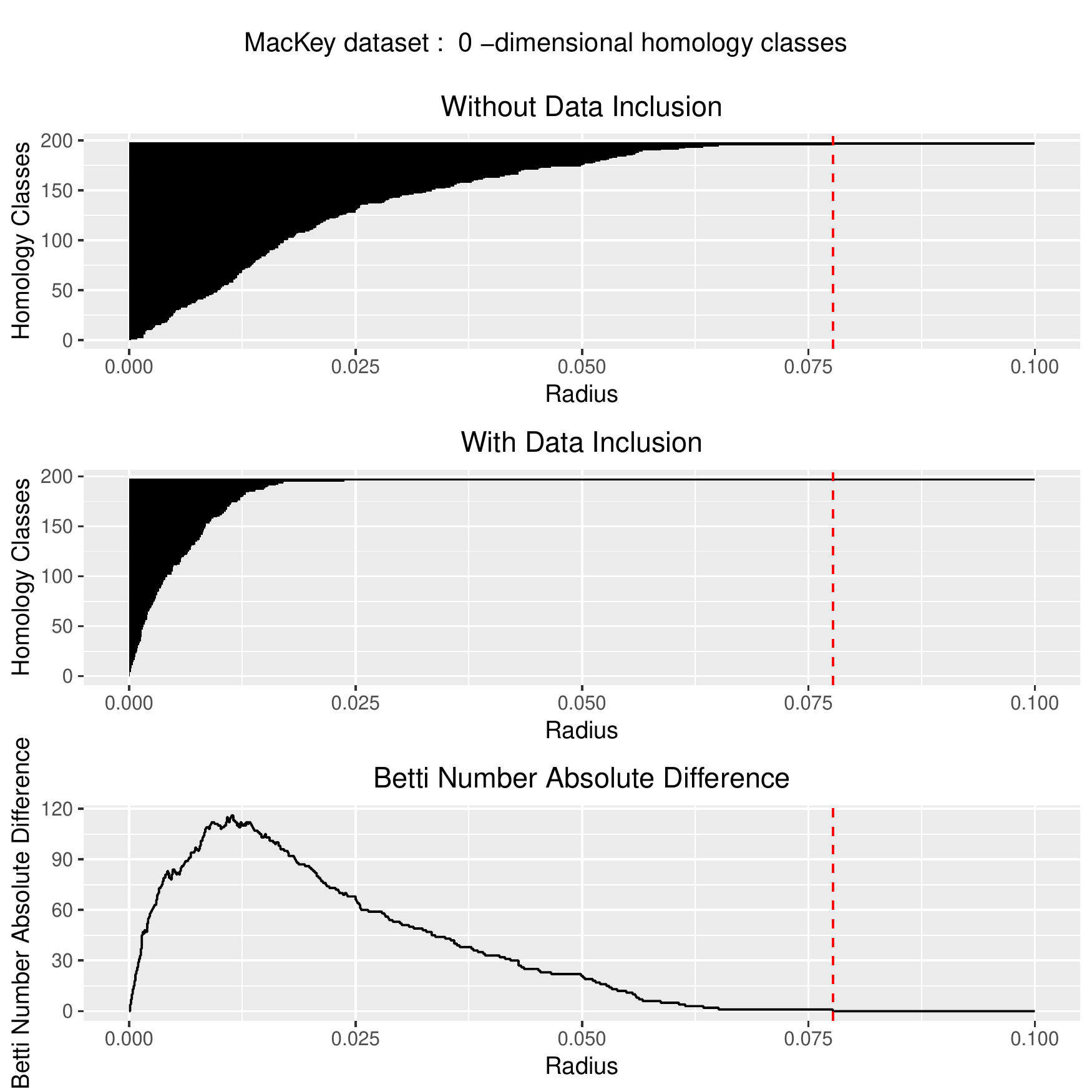}
\caption{Graphs produced from $0$-dimensional homology classes which correspond to, from top to bottom: i) Barcode plots generated using the Mackey-Glass attractor experiment with $500$ samples; ii) Barcode plots generated using the perturbed dataset with $1,000$ samples; and, finally, iii) the values for the generalization measure $|\beta_0(\mathcal{X}_{r_i}) - \beta_0(\mathcal{X}_{r_i}^m)|$. The red-dashed lines marks the initial value of the intervals which ensure $|\beta_0(\mathcal{X}_{r_i}) - \beta_0(\mathcal{X}_{r_i}^m)| = 0$.}
\label{fig:mackey-crit-0}
\end{figure}

\newpage
\begin{figure}[htb!]
\centering
\includegraphics[scale=0.64]{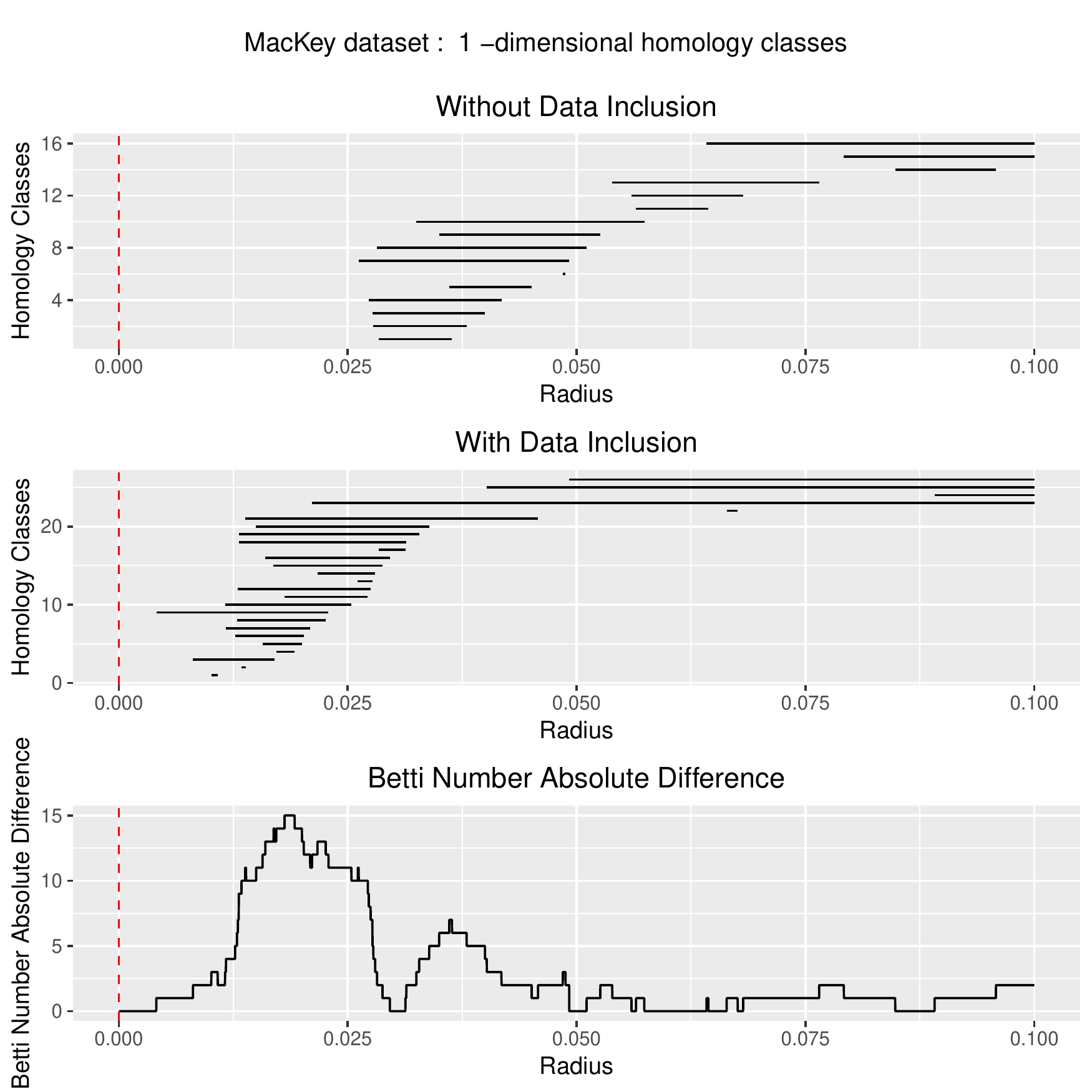}
\caption{Graphs produced from one-dimensional homology classes which correspond to, from top to bottom: i) Barcode plots generated using the Mackey-Glass attractor prior described with $500$ samples; ii) Barcode plots generated using the perturbed dataset with $1,000$ samples; and, finally, iii) the values for the generalization measure $|\beta_1(\mathcal{X}_{r_i}) - \beta_1(\mathcal{X}_{r_i}^m)|$. The red-dashed lines marks the initial value of the intervals which ensure $|\beta_1(\mathcal{X}_{r_i}) - \beta_1(\mathcal{X}_{r_i}^m)| = 0$.}
\label{fig:mackey-crit-1}
\end{figure}

\end{document}